\pdfoutput=1
\documentclass[11pt]{article}

\usepackage[]{emnlp2021}

\usepackage{times}
\usepackage{latexsym}

\usepackage[T1]{fontenc}

\usepackage[utf8]{inputenc}

\usepackage{microtype}

\usepackage{array}
\usepackage{color}
\usepackage{bm}
\usepackage{enumitem}
\usepackage{booktabs}
\usepackage{amssymb}
\usepackage{amsfonts}
\usepackage{amsmath}
\usepackage{multirow}
\usepackage{graphicx}
\usepackage{url}
\usepackage{footmisc}

\makeatletter
\let\MYcaption\@makecaption
\makeatother

\usepackage{subcaption}
\makeatletter
\let\@makecaption\MYcaption
\makeatother

\newcommand{\todo}[1]{}
\renewcommand{\todo}[1]{{\color{red} TODO: {#1}}}
\newcommand{\modified}[1]{}
\renewcommand{\modified}[1]{{\color{purple} MOD: {#1}}}
\newcommand{\kiyono}[1]{}
\renewcommand{\kiyono}[1]{{\color{purple} Kiyono: {#1}}}
\newcommand{\js}[1]{}
\renewcommand{\js}[1]{{\color{purple} JS: {#1}}}
\newcommand{\sosk}[1]{}
\renewcommand{\sosk}[1]{{\color{purple} Sosk: {#1}}}

\newcommand{\modeldim}{\ensuremath{\mathbb{R}^{D}}}
\newcommand{\baseline}{APE}
\newcommand{\relative}{RPE}
\newcommand{\proposed}{SHAPE}
\newcommand{\fulldata}{\textsc{Vanilla}}
\newcommand{\shortdata}{\textsc{Extrapolate}}
\newcommand{\longdata}{\textsc{Interpolate}}
\newcommand{\sep}{$ \langle \textit{sep} \rangle $}

\title{SHAPE: Shifted Absolute Position Embedding for Transformers}

\author{
Shun Kiyono$^{\spadesuit,\heartsuit}$\hspace{1em}
Sosuke Kobayashi$^{\heartsuit,\diamondsuit}$\hspace{1em}
Jun Suzuki$^{\heartsuit,\spadesuit}$\hspace{1em}
Kentaro Inui$^{\heartsuit,\spadesuit}$\\[2pt]
$^{\spadesuit}$ RIKEN\hspace{1em}
$^{\heartsuit}$ Tohoku University\hspace{1em}
$^{\diamondsuit}$ Preferred Networks, Inc.\hspace{1em} \\
\texttt{shun.kiyono@riken.jp},\hspace{1em}\texttt{sosk@preferred.jp},\\
\texttt{\{jun.suzuki, inui\}@tohoku.ac.jp}
}

\begin{document}
\maketitle
\begin{abstract}
  Position representation is crucial for building position-aware representations in Transformers. 
  Existing position representations suffer from a lack of generalization to test data with unseen lengths or high computational cost. %
  We investigate shifted absolute position embedding (SHAPE) to address both issues.
  The basic idea of SHAPE is to achieve \emph{shift invariance}, which is a key property of recent successful position representations, by randomly shifting absolute positions during training.
  We demonstrate that SHAPE is empirically comparable to its counterpart while being simpler and faster\footnote{The code is available at \url{https://github.com/butsugiri/shape}.}.
\end{abstract}

\section{Introduction}
Position representation plays a critical role in self-attention-based encoder-decoder models (Transformers)~\citep{vaswani:2017:NIPS},
enabling the self-attention to recognize the order of input sequences.
Position representations have two categories~\citep{dufter2021position}: absolute position embedding (APE)~\citep{gehring:2017:icml,vaswani:2017:NIPS} and relative position embedding (RPE)~\citep{shaw:2018:naacl}.
With APE, each position is represented by a unique embedding, which is added to inputs.
RPE represents the position based on the relative distance between two tokens in the self-attention mechanism.

\relative{} outperforms \baseline{} on sequence-to-sequence tasks~\cite{narang:2021:arxiv,neishi:2019:conll} due to \textit{extrapolation}, i.e., the ability to generalize to sequences that are longer than those observed during training~\citep{newman:2020:eos}.
\citet{wang:2021:bert-position} reported that one of the key properties contributing to RPE's superior performance is
\textit{shift invariance}\footnote{\textit{Shift invariance} is also known as \textit{translation invariance}.}, 
the property of a function to not change its output even if its input is shifted.
However, unlike APE, RPE's formulation strongly depends on the self-attention mechanism.
This motivated us to explore a way to incorporate the benefit of shift invariance in APE.

\begin{figure}[t]
  \centering 
  \includegraphics[width=\hsize]{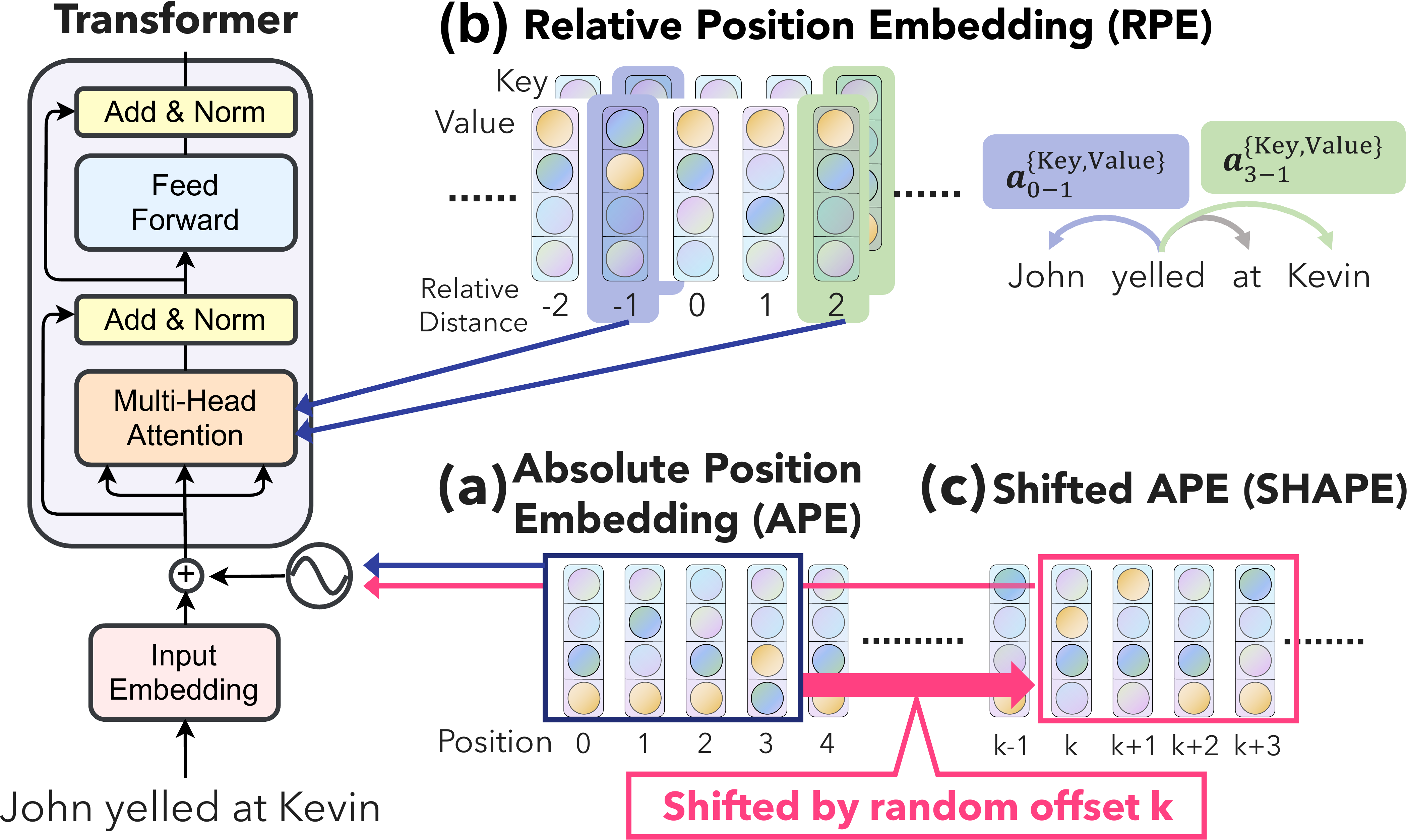}
   \caption{Overview of position representations. (a) \baseline{} and (c) \proposed{} consider \textit{absolute} positions in the \textit{input layer}, whereas (b) \relative{} considers the \textit{relative} position of a given token pair in the \textit{self-attention mechanism}.}
   \label{fig:overview}
\end{figure}

A promising approach to achieving shift invariance while using absolute positions is to randomly shift positions during training.
A similar idea can be seen in several contexts, e.g., computer vision \citep{Goodfellow-et-al-2016} and question-answering in NLP~\citep{geva-etal-2020-injecting}.
APE is no exception; a random shift should force Transformer to capture the relative positional information from absolute positions.
However, the effectiveness of a random shift for incorporating shift invariance in APE is yet to be demonstrated. 
Thus, we formulate APE with a random shift as a variant of position representation, namely, \emph{Shifted Absolute Position Embedding} (\emph{\proposed{}}; Figure~\ref{fig:overview}c), and conduct a thorough investigation.
In our experiments, we first confirm that Transformer with \proposed{} learns to be shift-invariant.
We then demonstrate that \proposed{} achieves a performance comparable to RPE in machine translation.
Finally, we reveal that Transformer equipped with shift invariance shows not only better extrapolation ability but also better \textit{interpolation} ability, i.e., it can better predict rare words at positions observed during the training. %

\section{Position Representations}
\label{section:position-representations}

Figure~\ref{fig:overview} gives an overview of the position representations compared in this paper.
We denote a source sequence $\bm{X}$ as a sequence of $I$ tokens, namely, $\bm{X}=(x_1,\dots,x_I)$.
Similarly, let $\bm{Y}$ represent a target sequence of $J$ tokens $\bm{Y}=(y_1,\dots,y_J)$.

\subsection{Absolute Position Embedding (APE)}
APE provides each position with a unique embedding (Figure~\ref{fig:overview}a).
Transformer with APE computes the input representation as the sum of the word embedding and the position embedding for each token $x_i \in \bm{X}$ and $y_j \in \bm{Y}$.

Sinusoidal positional encoding~\citep{vaswani:2017:NIPS} is a deterministic function of the position and the \textit{de facto} standard APE for Transformer\footnote{Learned position embedding~\citep{gehring:2017:icml} is yet another variant of APE; however, we exclusively focus on sinusoidal positional encoding as its performance is comparable~\citep{vaswani:2017:NIPS}.}.
Specifically, for the $i$-th token, the $m$-th element of position embedding $\mathrm{PE}(i, m)$ is defined as
\begin{align}
\mathrm{PE}(i, m) \!= \!
 \begin{cases}
   \label{eq:absolute_position}
   \sin \left(\frac{i}{10000^{\frac{2m}{D}}}\right)  & m \  \textrm{is even}\\
   \cos \left(\frac{i}{10000^{\frac{2m}{D}}}\right) & m \ \textrm{is odd} 
  \end{cases},
\end{align}
where $D$ denotes the model dimension.

\subsection{Relative Position Embedding (RPE)}
\label{sec:rpe}

RPE~\citep{shaw:2018:naacl} incorporates position information by considering the relative distance between two tokens in the self-attention mechanism (Figure~\ref{fig:overview}b).
For example, \citet{shaw:2018:naacl} represent the relative distance between the $i$-th and $j$-th tokens with relative position embeddings $\bm{a}_{i-j}^{\mathrm{Key}}, \bm{a}_{i-j}^{\mathrm{Value}} \in \modeldim{}$.
These embeddings are then added to key and value representations, respectively.

RPE outperforms APE on out-of-distribution data in terms of sequence length owing to its innate \textit{shift invariance}~\citep{rosendahl:2019:iwslt,neishi:2019:conll,narang:2021:arxiv,wang:2021:bert-position}.
However, the self-attention mechanism of RPE involves more computation than that of APE\footnote{\citet{narang:2021:arxiv} reported that Transformer with RPE is up to 25\% slower than that with APE.}.
In addition, more importantly, RPE requires the modification of the architecture, while APE does not.
Specifically, RPE strongly depends on the self-attention mechanism; thus, it is not necessarily compatible with studies that attempt to replace the self-attention with a more lightweight alternative~\citep{kitaev2020reformer,choromanski:2020:performer,tay2020efficient}.

RPE, which was originally proposed by \citet{shaw:2018:naacl}, has many variants in the literature~\citep{dai:2019:transformer-xl,raffel:2020:t5,huang-etal-2020-improve,wang:2021:bert-position,wu-etal-2021-da}.
They aim to improve the empirical performance or the computational speed compared with the original RPE.
However, the original RPE is still a strong method in terms of the performance.
\citet{narang:2021:arxiv} conducted a thorough comparison on multiple sequence-to-sequence tasks and reported that the performance of the original RPE is comparable to or sometimes better than its variants.
Thus, we exclusively use the original RPE in our experiments.

\subsection{Shifted Absolute Position Embedding (SHAPE)}
\label{sec:proposed_method}
Given the drawbacks of RPE, we investigate \proposed{} (Figure~\ref{fig:overview}c) as a way to equip Transformer with shift invariance without any architecture modification or computational overhead on APE.
During training, \proposed{} shifts every position index of APE by a random offset.
This prevents the model from using absolute positions to learn the task and instead encourages the use of relative positions, which we expect to eventually lead to the learning of shift invariance.

Let $k$ represent an offset drawn from a discrete uniform distribution $\mathcal{U}\{0, K\}$ for each sequence and for every iteration during training, where $K \in \mathbb{N}$ is the maximum shift.
SHAPE only replaces $\mathrm{PE}(i, m)$ of APE in Equation~\ref{eq:absolute_position} with
\begin{align}
\mathrm{PE}(i+k, m). %
\label{eq:noisy-ape}
\end{align}
We independently sample $k$ for the source and target sequence.
\proposed{} can thus be incorporated into any model using APE with virtually no computational overhead since only the input is modified. 
Note that \proposed{} is equivalent to the original APE if we set $K=0$; in fact, we set $K=0$ during inference.
Thus, \proposed{} can be seen as a natural extension to incorporate shift invariance in \baseline{}. 

\proposed{} can be interpreted in multiple viewpoints.
For example, \proposed{} can be seen as a regularizer that prevents Transformer from overfitting to the absolute position; such overfitting is undesirable not only for extrapolation~\citep{neishi:2019:conll} but also for APE with length constraints~\citep{takase:2019:naacl,oka:2020:coling,oka:2021:using}.
In addition, \proposed{} can be seen as a data augmentation method because the randomly sampled $k$ shifts each instance into different subspaces during training.

\section{Experiments}
\label{sec:experiment}
Using machine translation benchmark data, we first confirmed that Transformer trained with \proposed{} learns shift invariance (Section~\ref{subsec:exp-check-invariance}).
Then, we compared \proposed{} with \baseline{} and \relative{} to investigate its effectiveness (Section~\ref{sec:result}).

\subsection{Experimental Configuration}
\label{subsec:experimental-configuration}

\noindent\textbf{Dataset}\hspace*{3mm} 
We used the WMT 2016 English-German dataset for training and followed \citet{ott:2018:scaling} for tokenization and subword segmentation~\citep{sennrich:2016:subword}.
We used newstest2010-2013 and newstest2014-2016 as the validation and test sets, respectively.

Our experiments consist of the following three distinct dataset settings:

\noindent\textbf{(i) \fulldata{}}:\hspace*{3mm} Identical to previous studies~\citep{vaswani:2017:NIPS,ott:2018:scaling}.

\noindent\textbf{(ii) \shortdata{}}:\hspace*{3mm}
  Shift-invariant models are typically evaluated in terms of extrapolation ability~\citep{wang:2021:bert-position,newman:2020:eos}.
  We replicated the settings of \citet{neishi:2019:conll}; the training set excludes pairs whose source or target sequence exceeds 50 subwords, while the validation and test sets are identical to \fulldata{}.

\noindent\textbf{(iii) \longdata{}}:\hspace*{3mm}
  We also evaluate the models from the viewpoint of \emph{interpolation}, which we define as the ability to generate tokens whose lengths are seen during training.
  Specifically, we evaluate interpolation using long sequences since, first, the generation of long sequences is an important research topic in NLP~\citep{bigbird:2021:nips,docmt:2021:survey}
  and second, in datasets with long sequences, the position distribution of each token becomes increasingly sparse. 
  In other words, tokens in the validation and test sets become unlikely to be observed in the training set at corresponding positions; we expect that shift invariance is crucial for addressing such position sparsity.

  In this study, we artificially generate a long sequence by simply concatenating independent sentences in parallel corpus.
  Specifically, given ten neighboring sentences of \fulldata{}, i.e., $\bm{X}_1, \dots, \bm{X}_{10}$ and $\bm{Y}_1, \dots, \bm{Y}_{10}$, we concatenate each sentence with a unique token \sep{}. 
  We also apply the same operation to the validation and test sets.

\noindent\textbf{Evaluation}\hspace*{3mm} 
We evaluate the performance with sacreBLEU~\citep{post:2018:sacrebleu}.
Throughout the experiment, we apply the moses detokenizer to the system output and then compute the \textit{detokenized} BLEU\footnote{Details of datasets and evaluation are in Appendix~\ref{appendix:dataset-summary}.}.

\noindent\textbf{Models}\hspace*{3mm} 
We adopt \textit{transformer-base}~\citep{vaswani:2017:NIPS} with \baseline{}, \proposed{}, or \relative{}, respectively.
Our implementations are based on OpenNMT-py~\citep{klein-etal-2017-opennmt}.
Unless otherwise stated, we use a fixed value ($K = 500$) for the maximum shift of \proposed{} to demonstrate that \proposed{} is robust against the choice of $K$.
We set the relative distance limit in \relative{} to 16 following \citet{shaw:2018:naacl} and \citet{neishi:2019:conll}\footnote{See Appendix~\ref{appendix:hyperparameter} for a list of hyperparameters.}.

\subsection{Experiment 1: Shift Invariance}
\label{subsec:exp-check-invariance}
We confirmed that \proposed{} learns shift invariance by comparing \baseline{} and \proposed{} trained on \longdata{}.

\noindent\textbf{Quantitative Evaluation: BLEU on Training Data}\hspace*{3mm}
We first evaluated if the model is robust to the order of sentences in each sequence.
We used the sub-sampled training data (10k pairs) of \longdata{} to eliminate the effect of unseen sentences; in this way, we can isolate the effect of sentence order.
Given a sequence in the original order (\textbf{Original}), $\bm{X}_1,\dots,\bm{X}_{10}$, we generated a \emph{swapped} sequence (\textbf{Swapped}) by moving the first sentence to the end, i.e., $\bm{X}_2,\dots,\bm{X}_{10},\bm{X}_1$.
The model then generates two sequences $\bm{Y}_1^{\prime}, \dots,\bm{Y}_{10}^{\prime}$ and $\bm{Y}_2^{\prime},\dots,\bm{Y}_{10}^{\prime}, \bm{Y}_{1}^{\prime}$.
Finally, we evaluated the BLEU score of $\bm{Y}_1^{\prime}$.
The result is shown in Table~\ref{tab:invariance-sanity-check}.
Here, \proposed{} has a much smaller performance drop than \baseline{} when evaluated on different sentence ordering.
This result indicates the shift invariance property of \proposed{}.

\begin{table}[t]
  \centering
  \tabcolsep 1.5mm
  \small
  \begin{tabular}{lcc|c}
  \toprule
              & \textbf{Original} & \textbf{Swapped} & \textbf{Performance Drop} \\ \midrule
  APE         & 28.81             & 20.74            & 8.07                      \\
  \proposed{} & 28.51             & 27.06            & 1.45                      \\ \bottomrule
  \end{tabular}
  \caption{BLEU score on the sub-sampled \textit{training} data of \longdata{} (10,000 pairs). In \textbf{Original} and \textbf{Swapped}, the order of input sequence is $\bm{X}_1,\dots,\bm{X}_{10}$ and $\bm{X}_2,\dots,\bm{X}_{10},\bm{X}_1$, respectively.}
  \label{tab:invariance-sanity-check}
\end{table}

\begin{figure}[t]
  \centering 
  \includegraphics[width=\hsize]{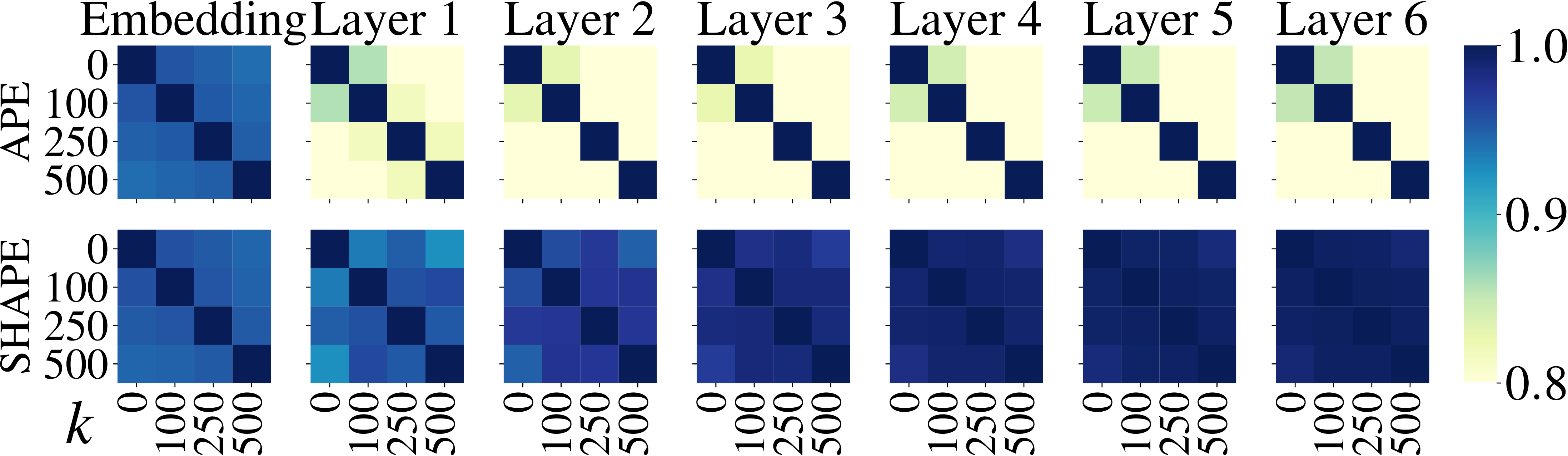}
   \caption{Cosine similarities of the encoder hidden states with different offsets $k \in \{0, 100, 250, 500\}$. Only the representation of \proposed{} is invariant with $k$.}
   \label{fig:ape-vs-noisy-ape-heatmap}
\end{figure}

\noindent\textbf{Qualitative Evaluation: Similarities of Representations}\hspace*{3mm} We also qualitatively confirmed the shift invariance as shown in Figure~\ref{fig:ape-vs-noisy-ape-heatmap}.
The figure illustrates how the offset $k$ changes the encoder representations of trained models \baseline{} and \proposed{}.
Given the two models and an input sequence $\bm{X}$, we computed the encoder hidden states of the given input sequence for each  $k \in \{0, 100, 250, 500\}$. %
For each position $i$, we computed the cosine similarity ($\mathrm{sim}$) of the hidden states from two offsets, i.e., $\bm{h}_i^{k_1}, \bm{h}_i^{k_2} \in \modeldim{}$, and computed its average across the positions as
\begin{align}
\frac{1}{I}\sum_{i=1}^{I} \mathrm{sim}(\bm{h}_{i}^{k_1}, \bm{h}_{i}^{k_2}).
\end{align}
As shown in Figure~\ref{fig:ape-vs-noisy-ape-heatmap}, \proposed{} builds a shift-invariant representation; regardless of the offset $k$, the cosine similarity is almost always 1.0.
Such invariance is nontrivial because the similarity of \baseline{} does not show similar characteristics\footnote{Additional figures are available in Appendix~\ref{appendix:more-figures}.}.

\begin{table}[t]
  \centering
  \small
  \begin{tabular}{llcc|c}
  \toprule
  \multicolumn{1}{c}{\textbf{Dataset}} & \multicolumn{1}{c}{\textbf{Model}} & \textbf{Valid} & \textbf{Test}  & \textbf{Speed} \\ \midrule
  \fulldata{}      & \baseline{}${}^{\dag}$               & 23.61 & 30.46 & x1.00  \\
                   & \relative{}${}^{\dag}$               & 23.67 & 30.54 & x0.91  \\
                   & \proposed{}${}^{\dag}$               & 23.63 & 30.49 & x1.01  \\
  \midrule
  \shortdata{}    & \baseline{}               & 22.18 & 29.22 & x1.00  \\
                  & \relative{}               & 22.97 & 29.86 & x0.91  \\
                  & \proposed{}               & 22.96 & 29.80 & x0.99  \\
  \midrule
  \longdata{}     & \baseline{}               & 31.40 & 38.23 & x1.00  \\
                  & \relative{}${}^{*}$               & -     & -     & -     \\
                  & \proposed{}               & 32.50 & 39.09 & x0.99  \\ \bottomrule
  \end{tabular}
  \caption{BLEU scores on newstest2010-2016. \textbf{Valid} is the average of newstest2010-2013. \textbf{Test} is the average of newstest2014-2016. The scores for individual newstests are available in Appendix~\ref{appendix:detailed-results}. $\dag$: the values are averages of five distinct trials with five different random seeds. $*$: not available as the implementation was very slow. \textbf{Speed} is the relative speed to \baseline{} (larger is faster).}
  \label{tab:bleu-score-aggregated}
\end{table}

\subsection{Experiment 2: Performance Comparison}
\label{sec:result}
\begin{figure}[t]
  \centering 
  \includegraphics[width=\hsize]{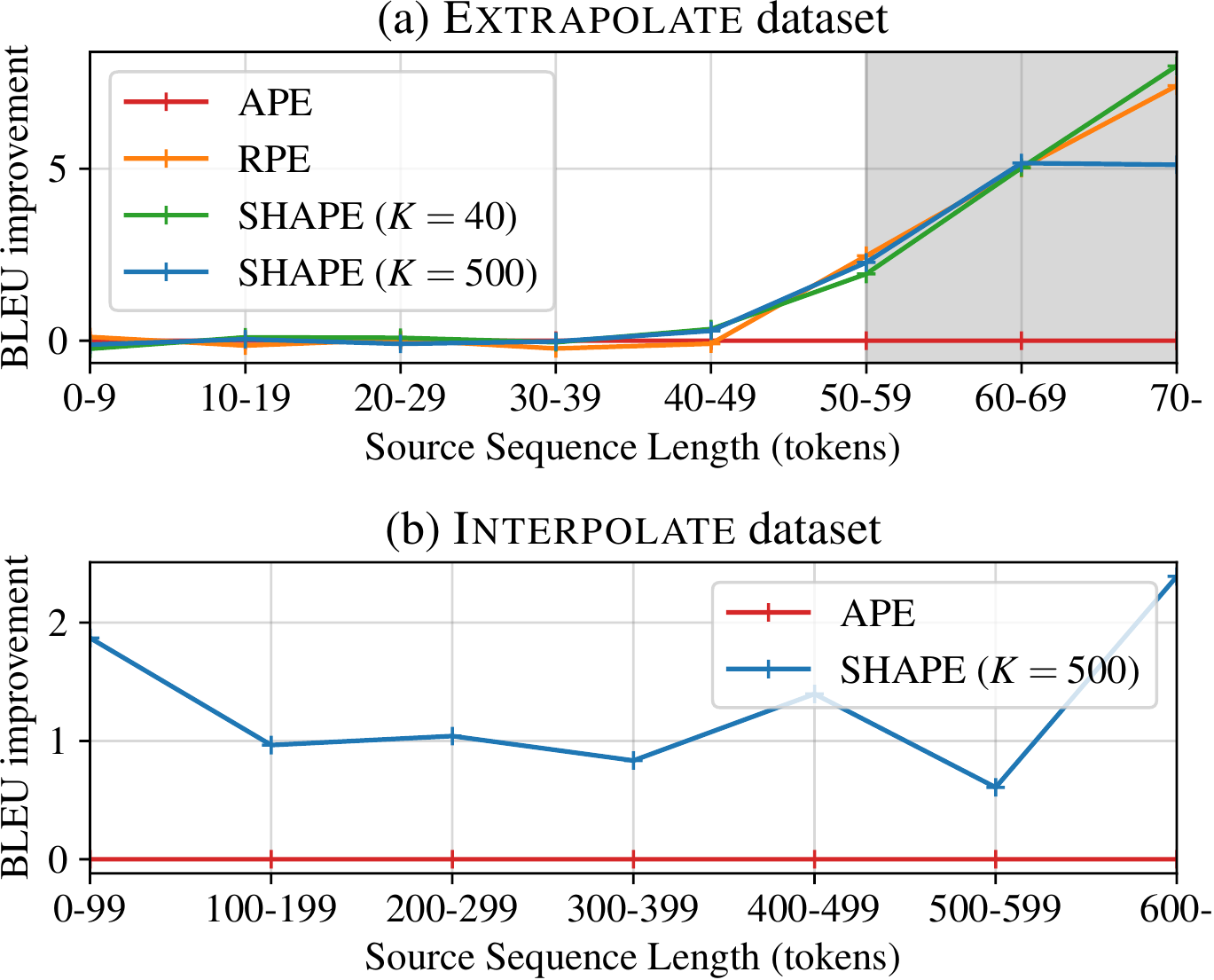}
   \caption{BLEU score improvement from \baseline{} on validation and test sets with respect to the source sequence length. The gray color means no training data.}
   \label{fig:bleu-by-length}
\end{figure}

We compared the overall performance of position representations on the validation and test sets as shown in Table~\ref{tab:bleu-score-aggregated}.
Figure~\ref{fig:bleu-by-length} shows the BLEU improvement of \relative{} and \proposed{} from \baseline{} with respect to the source sequence length\footnote{The same graph with absolute BLEU is in Appendix~\ref{appendix:detailed-results}.}.

On \textbf{\fulldata{}}, the three models show comparable results.
\baseline{} being comparable to \relative{} is inconsistent with the result reported by \citet{shaw:2018:naacl}; we assume that this is due to a difference in implementation.
In fact, \citet{narang:2021:arxiv} have recently reported that improvements in Transformer often do not transfer across implementations.

On \textbf{\shortdata{}}, \relative{} (29.86) outperforms \baseline{} (29.22) by approximately 0.6 BLEU points on the test set; this is consistent with the result reported by \citet{neishi:2019:conll}.
Moreover, \proposed{} achieves comparable test performance to \relative{} (29.80).
According to Figure~\ref{fig:bleu-by-length}a, both \relative{} and \proposed{} have improved extrapolation ability, i.e., better BLEU scores on sequences longer than those observed during training.
In addition, Figure~\ref{fig:bleu-by-length}a shows the performance of \proposed{} with the maximum shift $K=40$ that was chosen on the basis of the BLEU score for the validation set.
This model outperforms \relative{}, achieving  BLEU scores of 23.12 and 29.86 on the validation and test sets, respectively.
These results indicate that \proposed{} can be a better alternative to \relative{}.

On \textbf{\longdata{}}, we were unable to train \relative{} because its training was prohibitively slow\footnote{A single gradient step of \relative{} took about 5 seconds, which was 20 times longer than that of \baseline{} and \proposed{}. We assume that the RPE implementation available in OpenNMT-py has difficulty in dealing with long sequences.\label{fn:latency}}.
Similarly to \shortdata{}, \proposed{} (39.09) outperforms \baseline{} (38.23) on the test set.
Figure~\ref{fig:bleu-by-length}b shows that \proposed{} consistently outperformed \baseline{} for every sequence length. %
From this result, we find that the shift invariance also improves the \emph{interpolation ability} of Transformer. %

\begin{figure}[t]
  \begin{subfigure}[b]{\hsize}
    \centering
  \includegraphics[width=\hsize]{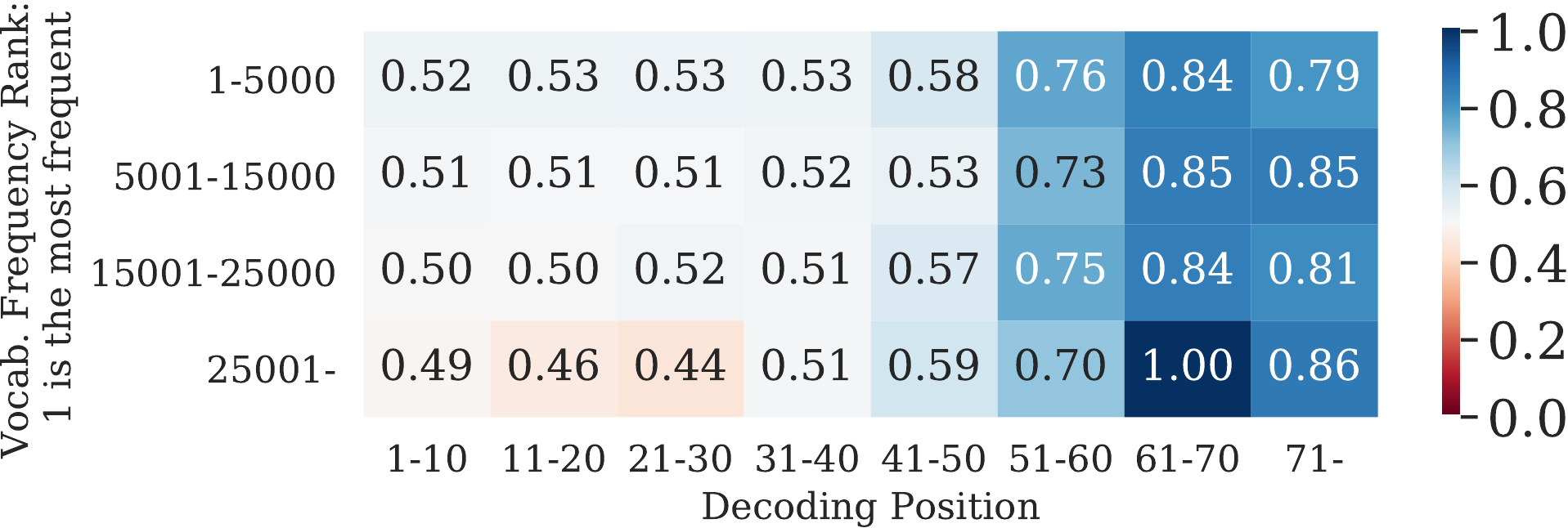}
   \caption{\shortdata{} dataset}
   \label{fig:position-1to50}
  \end{subfigure}
  \begin{subfigure}[b]{\hsize}
    \centering
  \includegraphics[width=\hsize]{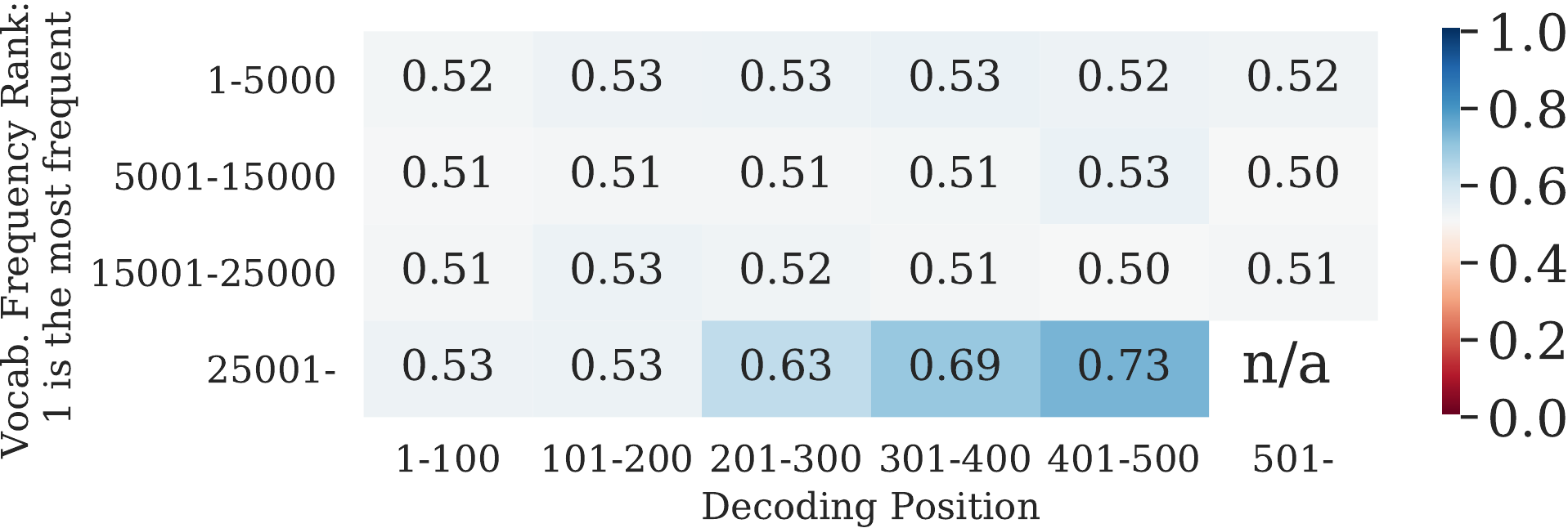}
   \caption{\longdata{} dataset}
   \label{fig:position-merge10}
  \end{subfigure}
  \caption{Tokenwise analysis on gold references: the value in each cell represents the ratio that \proposed{} assigns a higher score to a gold token than \baseline{}.}
   \label{fig:position-wise-analysis}
\end{figure}

\section{Analysis}
\label{sec:analysis}
This section provides a deeper analysis of how the model with translation invariance improves the performance.
We hereinafter exclusively focus on \baseline{} and \proposed{} because \proposed{} achieves comparable performance to \relative{}, and we were unable to train \relative{} on the \longdata{} dataset as explained in footnote \ref{fn:latency}.

As discussed in Section~\ref{sec:result}, Figure~\ref{fig:bleu-by-length} demonstrated that \proposed{} outperformed \baseline{} in terms of BLEU score.
However, BLEU evaluates two concepts simultaneously, that is, the token precision via n-gram matching and the output length via the brevity penalty~\citep{papineni-etal-2002-bleu}.
Thus, the actual source of improvement remains unclear.
We hereby exclusively analyzed the precision of token prediction. %
Specifically, we computed tokenwise scores assigned for gold references, and we then compared them  across the models; 
given a sequence pair $(\bm{X}, \bm{Y})$ and a trained model, we computed a score (i.e., log probability) $s_j$ for each token $y_j$ in a teacher-forcing manner.
Here, a higher score to gold token means better model performance.
We used the validation set for comparison.

Figure~\ref{fig:position-wise-analysis} shows the ratio that \proposed{} assigns a higher score to a gold token than \baseline{}, compared across for each position of the decoder.

\noindent\textbf{Better extrapolation means better token precision}\hspace*{3mm}
Figure~\ref{fig:position-1to50} shows that \proposed{} outperforms \baseline{}, especially in the right part of the heat map.
This area corresponds to sequences longer than those observed during training.
This result indicates that better extrapolation in terms of BLEU score means better token precision.

\noindent\textbf{Interpolation is particularly effective for rare tokens}\hspace*{3mm}
As shown in Figure~\ref{fig:position-merge10}, \proposed{} consistently outperforms \baseline{} and the performance gap is especially significant in the low-frequency region (bottom part). %
This indicates that \proposed{} predicts rare words better than \baseline{}.
One plausible explanation for this observation is that \proposed{} carries out data augmentation in the sense that in each epoch, the same sequence pair is assigned a different position depending on the offset $k$. 
Rare words typically have sparse position distributions in training data and thus benefit from the extra position assignment during training.

\section{Conclusion}

We investigated \proposed{}, a simple variant of \baseline{} with shift invariance.
We demonstrated that \proposed{} is empirically comparable to \relative{} yet imposes almost no computational overhead on \baseline{}.
Our analysis revealed that \proposed{} is effective at extrapolation to unseen lengths and interpolating rare words.
\proposed{} can be incorporated into the existing codebase with a few lines of code and no risk of a performance drop from \baseline{}; thus, we expect \proposed{} to be used as a drop-in replacement for \baseline{} and \relative{}.

\section*{Acknowledgements}
We thank the anonymous reviewers for their insightful comments.
We thank Sho Takase for valuable discussions.
We thank Ana Brassard, Benjamin Heinzerling, Reina Akama, and Yuta Matsumoto for their valuable feedback.
The work of Jun Suzuki was supported by JST Moonshot R\&D Grant Number JPMJMS2011 (fundamental research) 
and JSPS KAKENHI Grant Number 19H04162 (empirical evaluation).

\bibliography{anthology,custom}
\bibliographystyle{acl_natbib}

\clearpage
\appendix

\section{Summary of Datasets}
\label{appendix:dataset-summary}

\begin{figure*}[ht]
  \begin{subfigure}[b]{0.33\textwidth}
    \centering
  \includegraphics[width=\hsize]{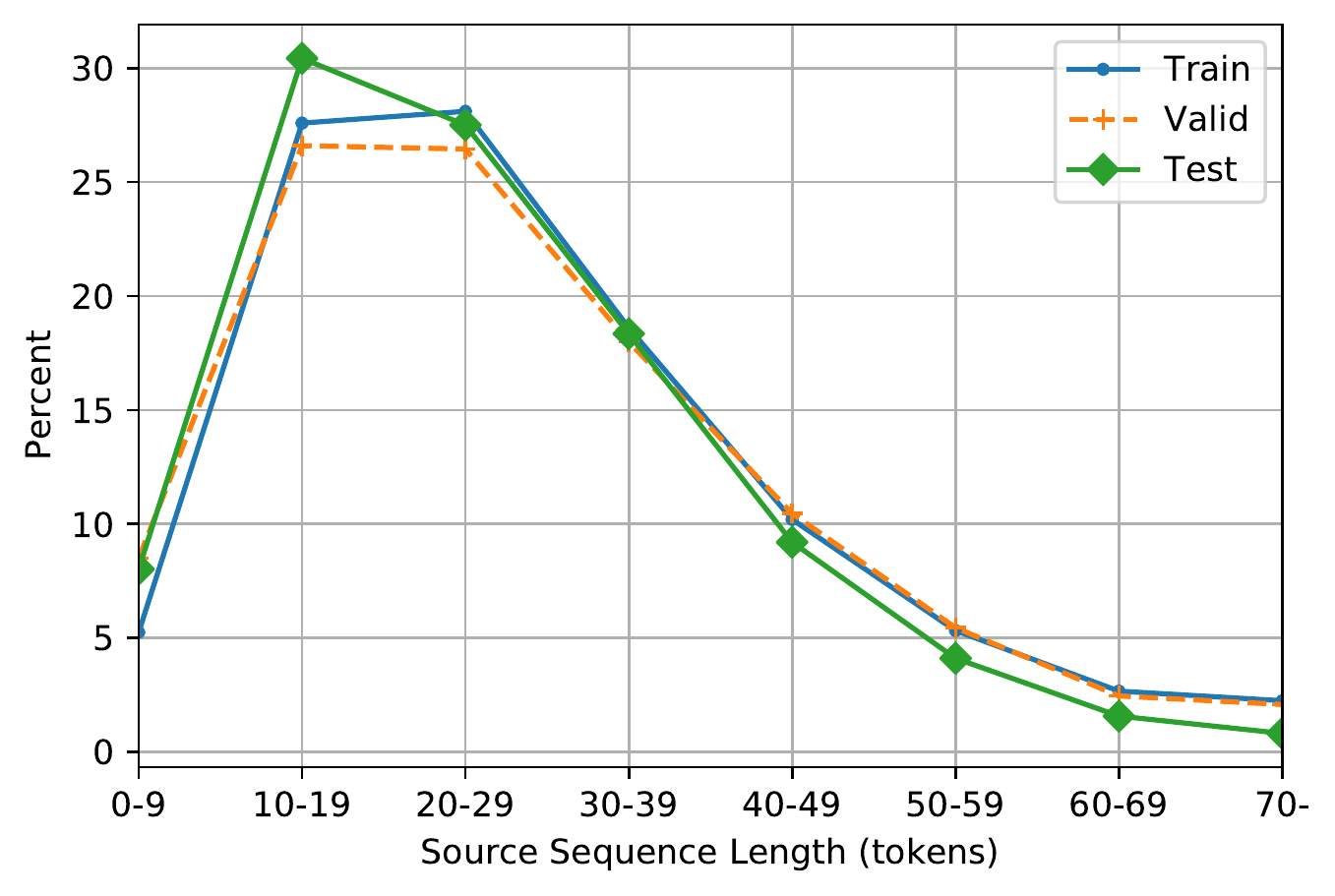}
   \caption{\fulldata{} dataset}
  \end{subfigure}
  \begin{subfigure}[b]{0.33\textwidth}
    \centering
  \includegraphics[width=\hsize]{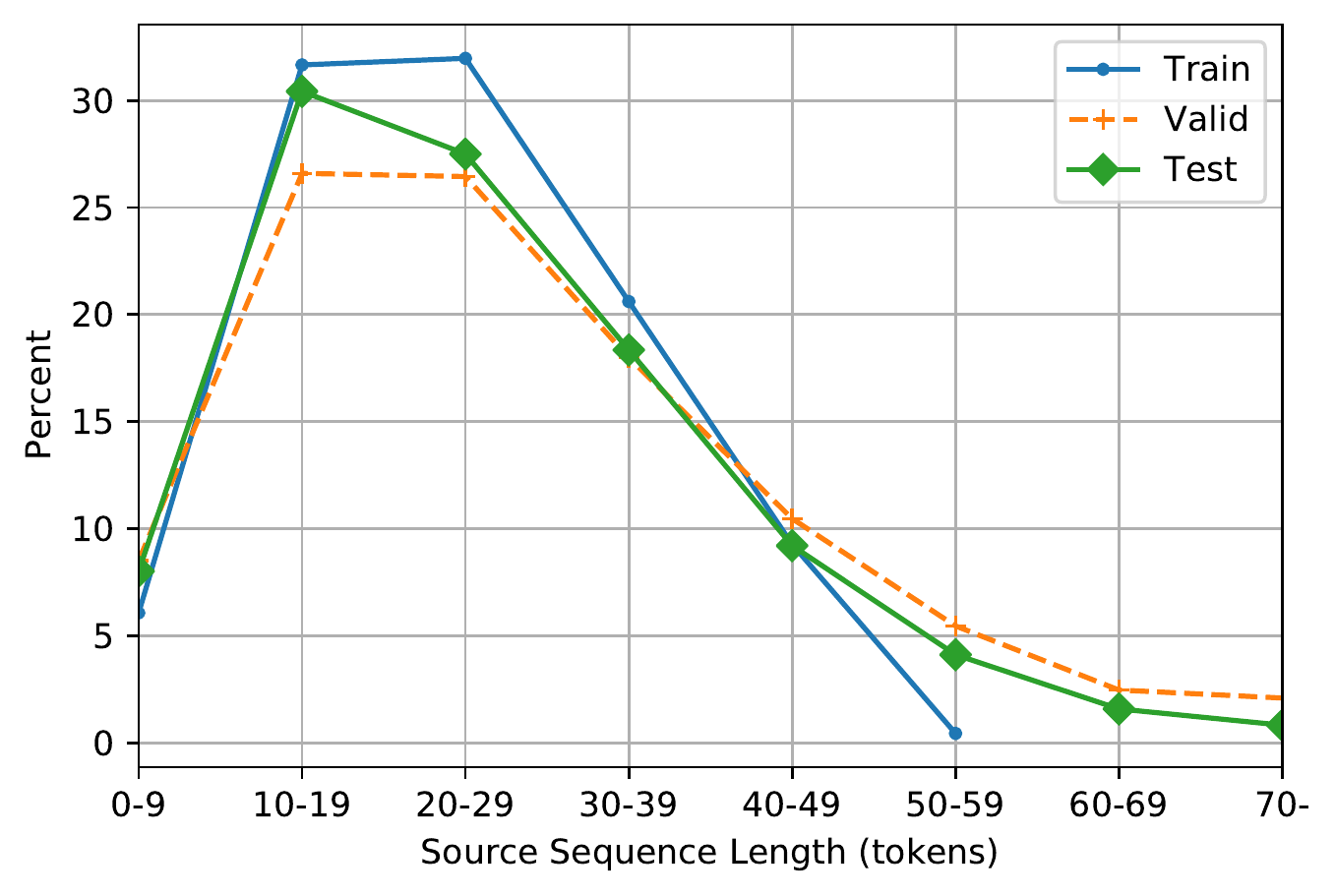}
   \caption{\shortdata{} dataset}
  \end{subfigure}
  \begin{subfigure}[b]{0.33\textwidth}
    \centering
  \includegraphics[width=\hsize]{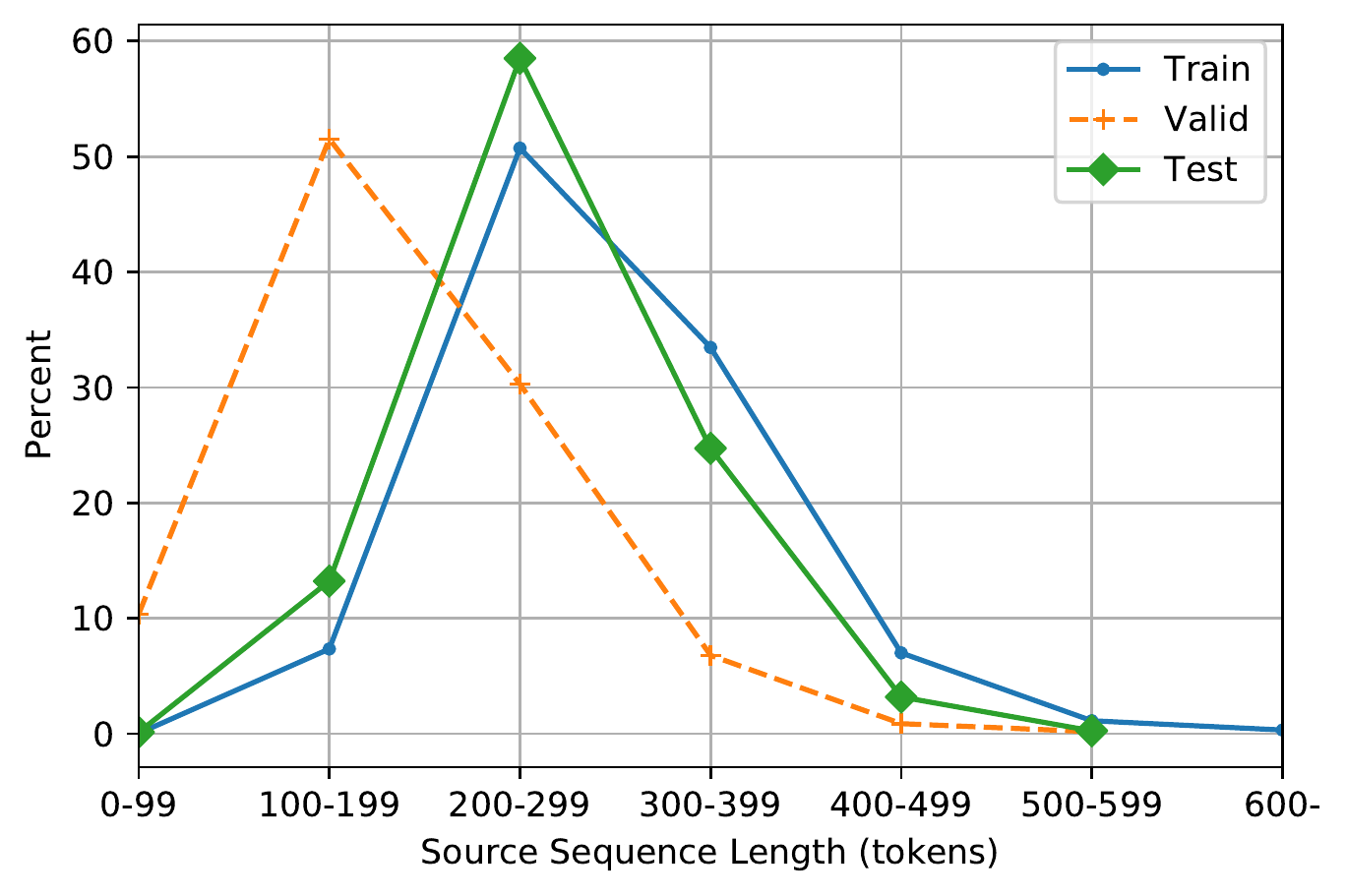}
   \caption{\longdata{} dataset}
  \end{subfigure}
  \caption{Distribution of source sequence length of each dataset.}
   \label{fig:length-distribution}
\end{figure*}

We summarized the statistics, preprocessing, and evaluation metrics of datasets used in our experiment in Table~\ref{tab:dataset-stats}.
The length statistics are in Figure~\ref{fig:length-distribution}.

\begin{table*}[t]
  \small
  \tabcolsep 1mm
  \centering
  \begin{tabular}{lp{40mm}p{12mm}p{24mm}p{24mm}p{24mm}}
  \toprule
  \textbf{Dataset Name} & \textbf{Training Data} & \textbf{\# of Sent. Pairs in Training Data}  & \textbf{Validation}        & \textbf{Test}              & \textbf{Evaluation Metric}              \\ \midrule
  \fulldata{}  & WMT 2016 English-German & 4.5M                                 & newstest2010-2013 & newsetst2014-2016 & detokenized BLEU via sacreBLEU \\\midrule
  \shortdata{} &
    WMT 2016 English-German. We removed sequence pairs if the length of the source or target sentence exceeds 50 subwords. &
    3.9M &
    newstest2010-2013 &
    newsetst2014-2016 &
    detokenized BLEU via sacreBLEU \\\midrule
  \longdata{} &
    WMT 2016 English-German. Given neighboring ten sentence of \fulldata{}, i.e., $\bm{X}_1, \dots, \bm{X}_{10}$ and $\bm{Y}_1, \dots, \bm{Y}_{10}$, we concatenate each sentence with a special token \sep{}. &
    450K &
    newstest2010-2013. We concatenated sentences as in training data. &
    newstest2014-2016. We concatenated sentences as in training data.&
    detokenized BLEU via sacreBLEU \\ \bottomrule
  \end{tabular}
  \caption{Summary of statistics, preprocessing, and evaluation metric of datasets used in our experiment.}
  \label{tab:dataset-stats}
\end{table*}

\section{Hyperparameters}
\label{appendix:hyperparameter}

We present the list of hyperparameters used in our experiments in Table~\ref{tab:hyper-parameter}.
Hyperparameters for training Transformer follow the recipe available in the official documentation page of OpenNMT-py\footnote{\url{https://opennmt.net/OpenNMT-py/FAQ.html\#how-do-i-use-the-transformer-model}}.

\begin{table*}[t]
  \centering
  \small
  \tabcolsep=2pt
 \begin{tabular}{lp{110mm}}
  \toprule
     \textbf{Configurations}         &   \textbf{Selected Value} \\ \midrule
     Encoder-Decoder Architecture           &  \textit{transformer-base}~\citep{vaswani:2017:NIPS} \\
     Optimizer              &   Adam  ($\beta_{1}=0.9, \beta_{2}=0.98, \epsilon=1\times10^{-8}$)  \\
     Learning Rate Schedule &   ``Noam'' scheduler described in \citep{vaswani:2017:NIPS}     \\
     Warmup Steps           &   8,000  \\
     Learning Rate Scaling Factor${}^{\dag}$      &   2  \\
     Dropout                &   0.1  \\
     Gradient Clipping      &   None  \\
     Beam Search Width      & 4        \\
     Label Smoothing        &   $\epsilon_{ls}=0.1$~\citep{szegedy:2016:rethinking}     \\
     Mini-batch Size        &   112k tokens  \\
     Number of Gradient Steps &   200,000  \\
     Averaging              &   Save checkpoint for every 5,000 steps and take an average of last 10 checkpoints \\
     Maximum Offset $K$ (for \proposed{})    &  We set $K=500$ for the most of the experiments. We manually tuned $K$ on validation BLEU for \shortdata{} from following range: \{10, 20, 30, 40, 100, 500\}, and report the score of $K=40$ in addition to $K=500$. We used a single random seed for the tuning.\\
     Relative Distance Limit (for \relative{})       & 16 following \citep{neishi:2019:conll}  \\
     GPU Hardware Used     &  DGX-1 and DGX-2     \\
  \bottomrule
  \end{tabular}
  \caption{List of hyperparameters. $\dag$: this corresponds to ``learning rate'' variable defined in OpenNMT-py framework.}
  \label{tab:hyper-parameter}
\end{table*}

\section{Similarities of Representations}
\label{appendix:more-figures}

\begin{figure*}[ht]
  \begin{subfigure}[b]{0.50\hsize}
    \centering
  \includegraphics[width=\hsize]{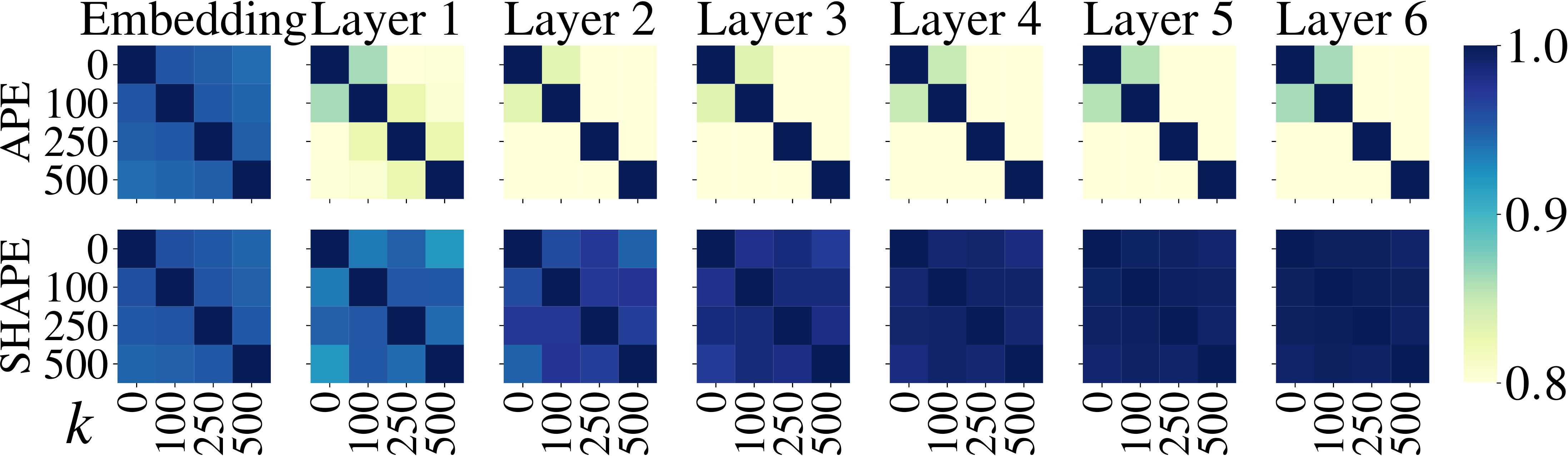}
   \caption{Sequence ID: \#1}
  \end{subfigure}
  \hfill
  \begin{subfigure}[b]{0.50\hsize}
    \centering
  \includegraphics[width=\hsize]{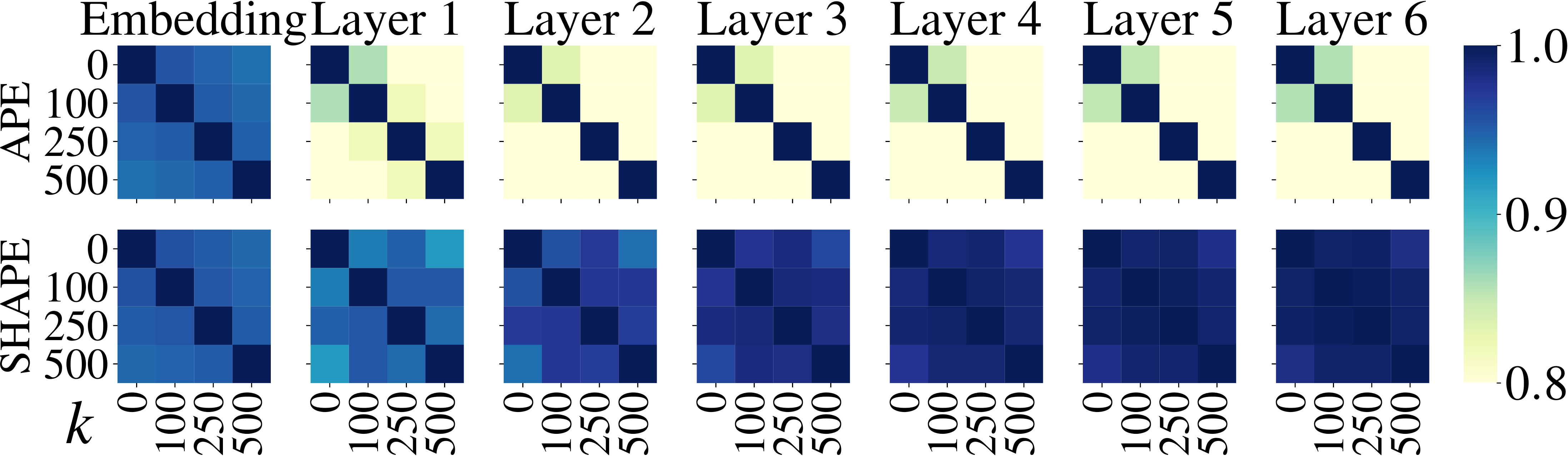}
   \caption{Sequence ID: \#2}
  \end{subfigure}
  \begin{subfigure}[b]{0.50\hsize}
    \centering
  \includegraphics[width=\hsize]{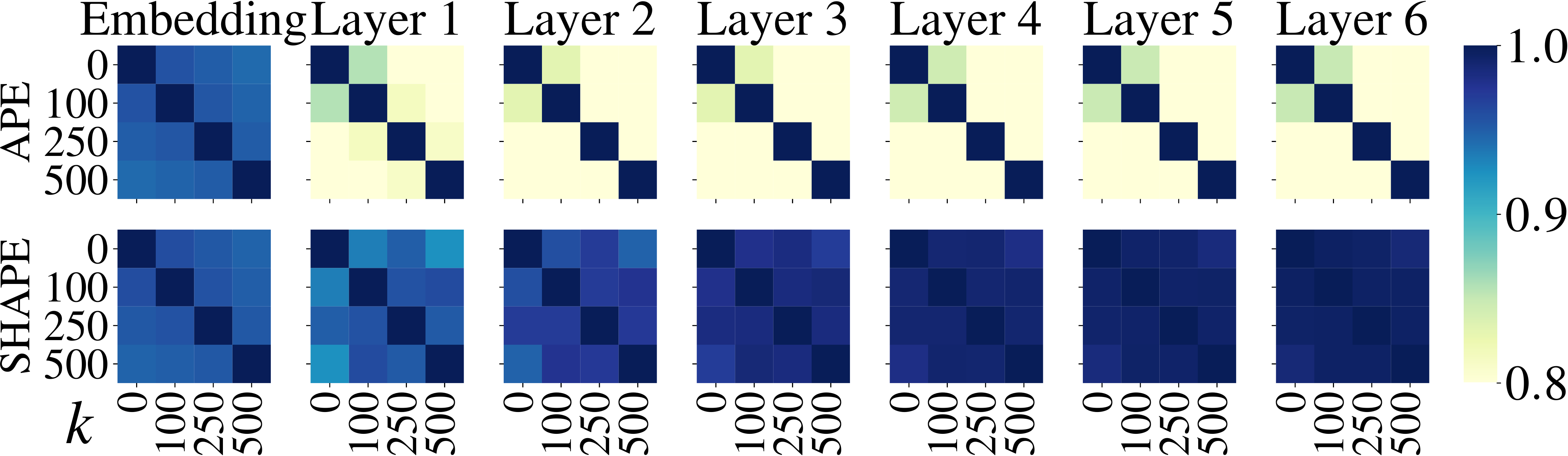}
   \caption{Sequence ID: \#3}
  \end{subfigure}
  \begin{subfigure}[b]{0.50\hsize}
    \centering
  \includegraphics[width=\hsize]{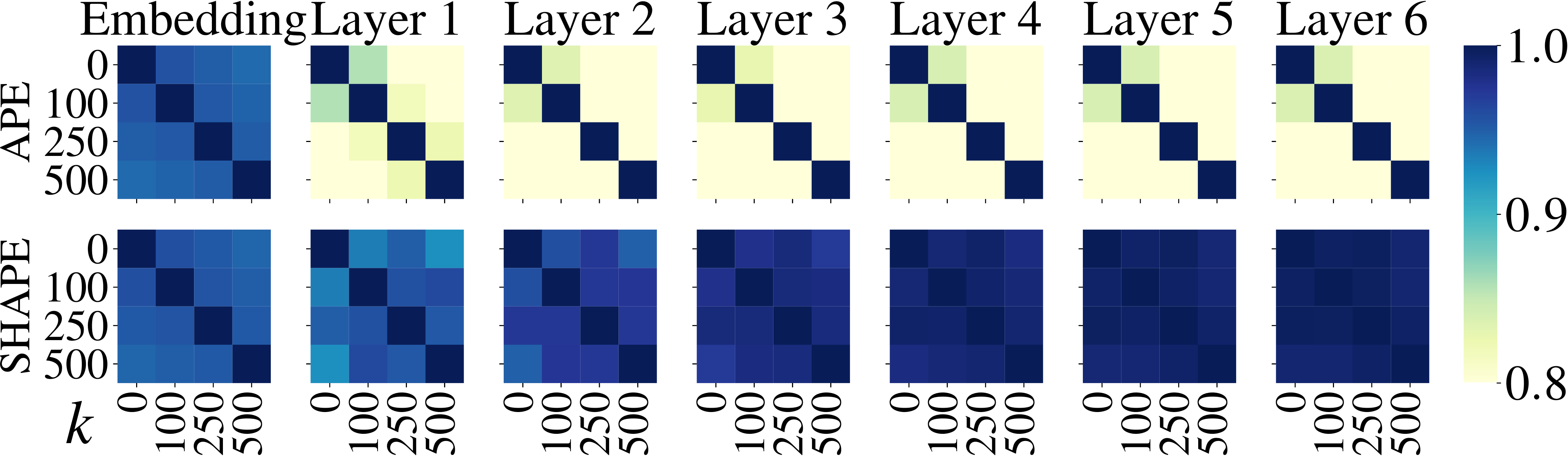}
   \caption{Sequence ID: \#4}
  \end{subfigure}
  \begin{subfigure}[b]{0.50\hsize}
    \centering
  \includegraphics[width=\hsize]{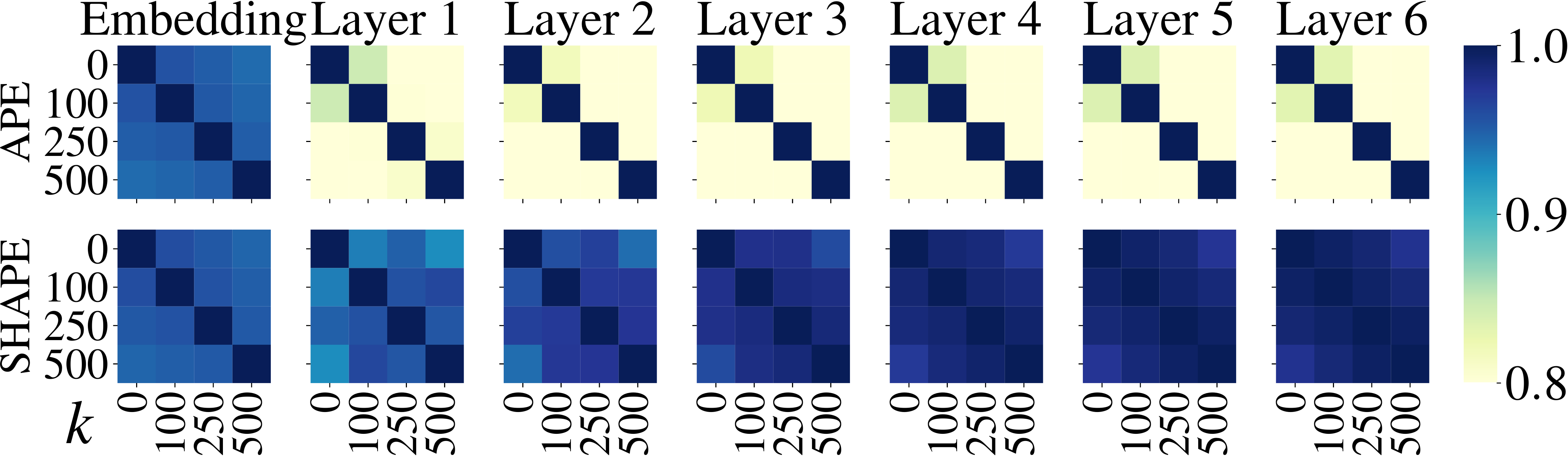}
   \caption{Sequence ID: \#5}
  \end{subfigure}
  \begin{subfigure}[b]{0.50\hsize}
    \centering
  \includegraphics[width=\hsize]{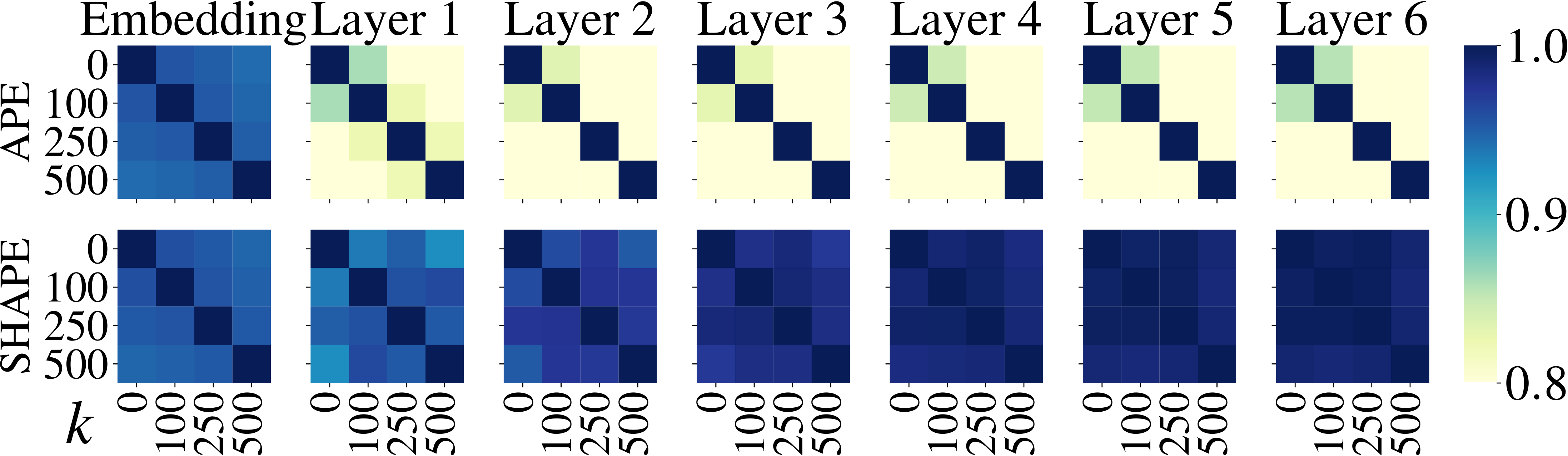}
   \caption{Sequence ID: \#6}
  \end{subfigure}
  \begin{subfigure}[b]{0.50\hsize}
    \centering
  \includegraphics[width=\hsize]{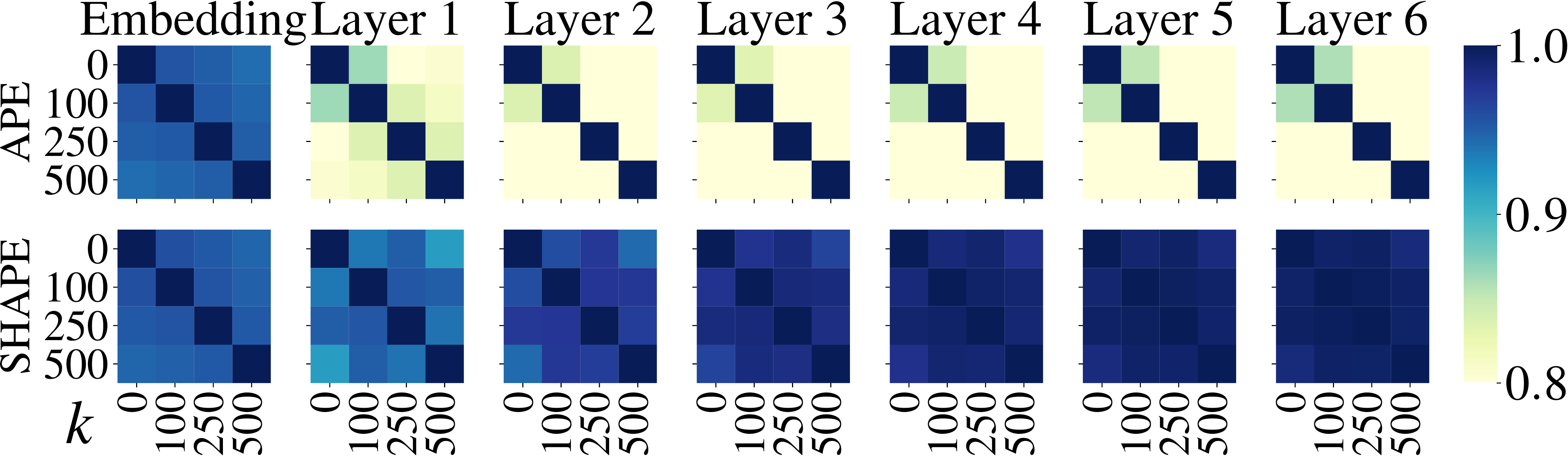}
   \caption{Sequence ID: \#7}
  \end{subfigure}
  \begin{subfigure}[b]{0.50\hsize}
    \centering
  \includegraphics[width=\hsize]{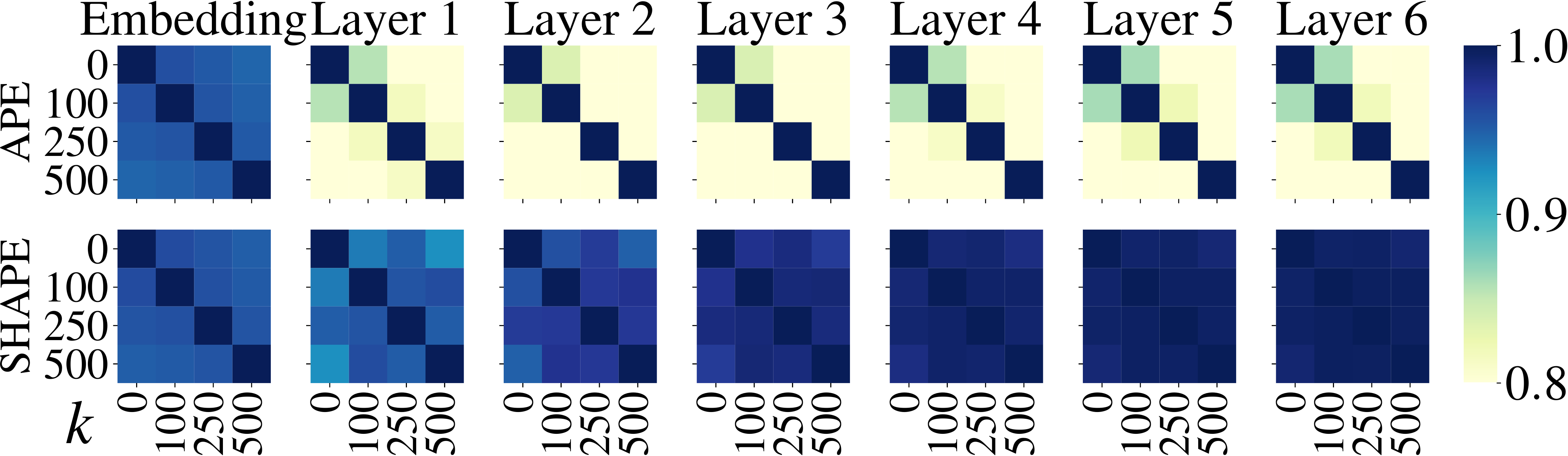}
   \caption{Sequence ID: \#8}
  \end{subfigure}
  \begin{subfigure}[b]{0.50\hsize}
    \centering
  \includegraphics[width=\hsize]{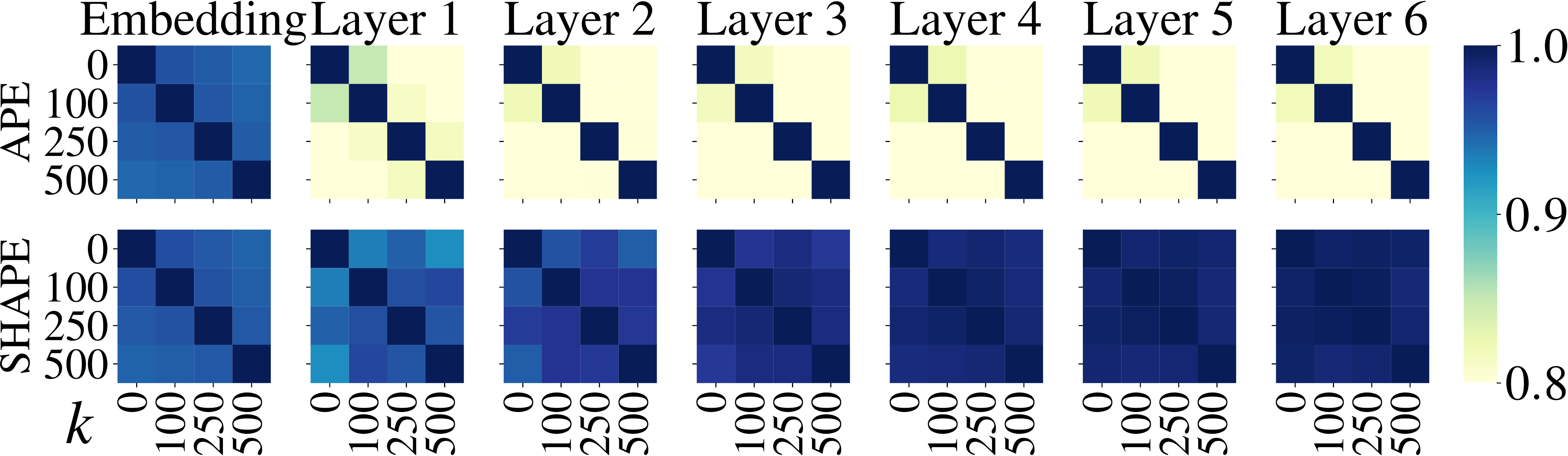}
   \caption{Sequence ID: \#9}
  \end{subfigure}
  \begin{subfigure}[b]{0.50\hsize}
    \centering
  \includegraphics[width=\hsize]{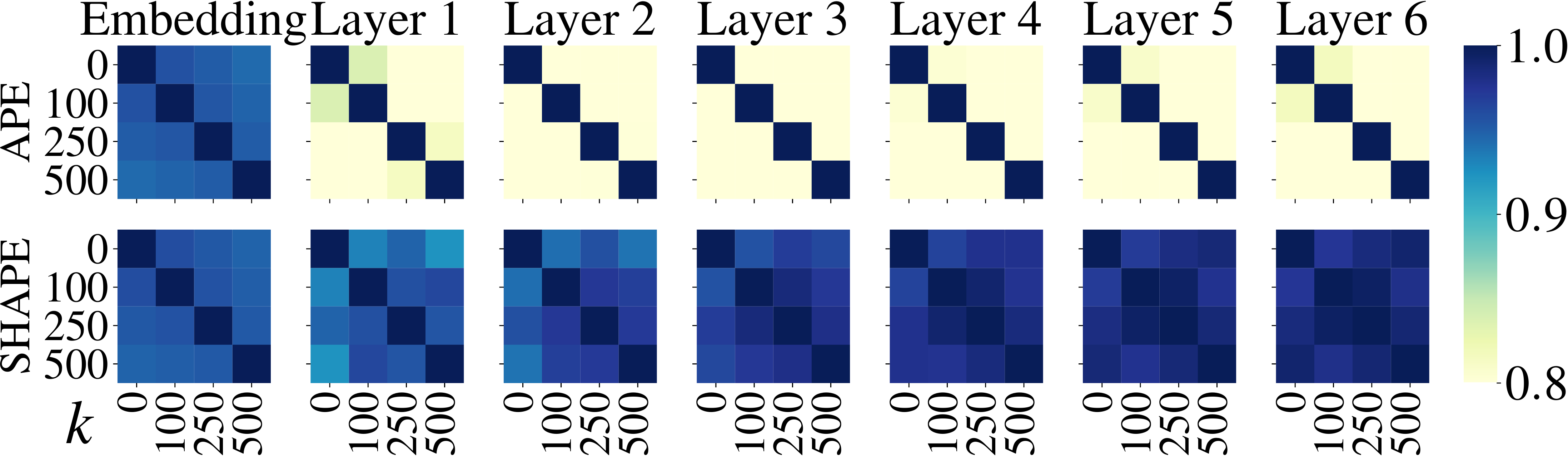}
   \caption{Sequence ID: \#10}
  \end{subfigure}

  \caption{Cosine similarities of encoder hidden states with different offsets $k \in \{0, 100, 250, 500\}$. Only the representation of \proposed{} is invariant with $k$.}
   \label{fig:additional-heatmap}
\end{figure*}

In Section~\ref{subsec:exp-check-invariance}, we presented Figure~\ref{fig:ape-vs-noisy-ape-heatmap} to qualitatively demonstrate that the representation of \proposed{} is shift-invariant.
We present ten additional figures that we created from ten additional instances in Figure~\ref{fig:additional-heatmap}.
The characteristic of the figures are consistent with that observed in Figure~\ref{fig:ape-vs-noisy-ape-heatmap}; the representation of \proposed{} is shift-invariant, whereas the representation of \baseline{} is not.

\section{Detailed BLEU Scores}
\label{appendix:detailed-results}

\begin{figure}[t]
  \centering 
  \includegraphics[width=\hsize]{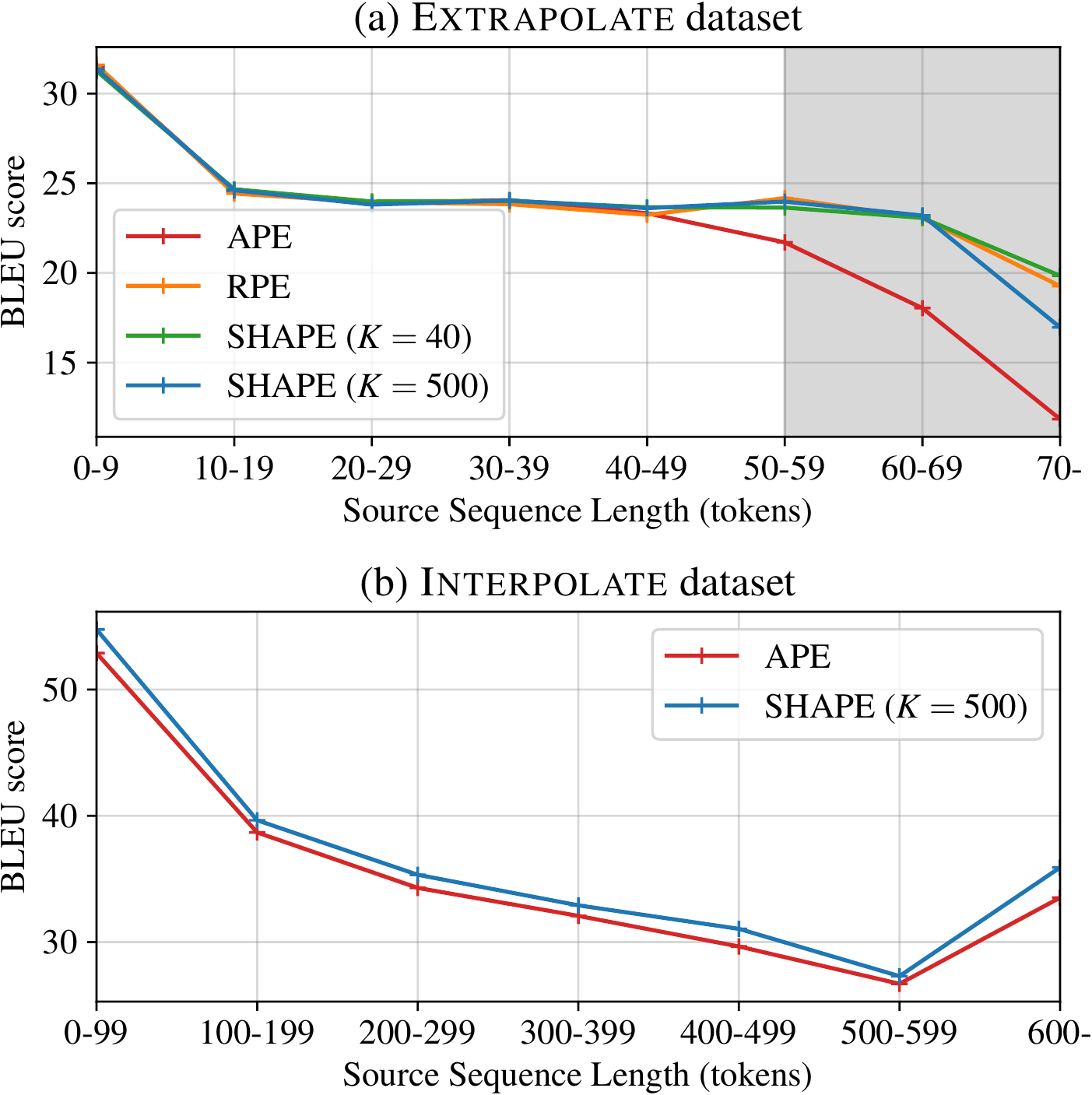}
   \caption{BLEU score on validation and test sets with respect to the source sequence length. The gray color means no training data.}
   \label{fig:bleu-by-length-absolute}
\end{figure}

We report the BLEU score on each of newstest2010-2016 in Table~\ref{tab:bleu-score}\footnote{SacreBLEU hash of \fulldata{} and \shortdata{} is: \texttt{BLEU+case.mixed+lang\\.en-de+numrefs.1+smooth.exp+\\test.wmt\{10,11,12,13,\\14/full,15,16\}+tok.13a+version.1.5.0}.}\footnote{SacreBLEU hash of \longdata{} is \texttt{BLEU+case.mixed+numrefs.1+smooth.exp+\\tok.13a+version.1.5.0}.}.
In addition, we report the performance of \baseline{}, \relative{}, and \proposed{} with respect to the source sequence lengths in Figure~\ref{fig:bleu-by-length-absolute}.

\begin{table*}[t]
  \centering
  \small
  \begin{tabular}{lccccccc|c|c}
  \toprule
  \multicolumn{1}{c}{\textbf{Model}} & \textbf{2010} & \textbf{2011} & \textbf{2012} & \textbf{2013} & \textbf{2014} & \textbf{2015} & \textbf{2016} & \textbf{Average} & \textbf{Speed} \\ \midrule
  \multicolumn{10}{c}{Dataset: \fulldata{}}                                                          \\\midrule
  APE${}^{\dag}$    & 24.22 & 21.98 & 22.20 & 26.06 & 26.95 & 29.98 & 34.46 & 26.55 & x1.00\\
  RPE${}^{\dag}$               & 24.29 & 22.05 & 22.22 & 26.13 & 27.00 & 30.00 & 34.61 & 26.61 & x0.91 \\
  \proposed{}${}^{\dag}$ & 24.18 & 22.01 & 22.23	& 26.08	& 26.89	& 30.12	& 34.48	& 26.57  & x1.01 \\ \midrule
  \multicolumn{10}{c}{Dataset: \shortdata{}}                                                         \\\midrule

  APE    & 22.69 & 20.36 & 20.72 & 24.94 & 26.24 & 28.79 & 32.62 & 25.19 & x1.00 \\
  RPE               & 23.46 & 21.19 & 21.69 & 25.54 & 26.80 & 29.43 & 33.34 & 25.92 & x0.91 \\
  \proposed{} & 23.60 & 21.24 & 21.53 & 25.45 & 26.54 & 29.22 & 33.63 & 25.89 & x0.99 \\\midrule
  \multicolumn{10}{c}{Dataset: \longdata{} ${}^{\ddag}$} \\\midrule
  APE    & 31.41 & 29.71 & 29.79 & 34.69 & 35.36 & 38.00 & 41.32 & 34.33 & x1.00 \\
  RPE${}^{*}$       & -     &  -     &   -    &  -     & -      &   -    & -      &  -  & -   \\
  \proposed{} & 32.71 & 30.77 & 30.96 & 35.54 & 35.72 & 39.18 & 42.37 &  35.32 & x0.99 \\ \bottomrule
  \end{tabular}
  \caption{BLEU scores on newstest2010-2016. \textbf{Average} column shows the macro average of all newstests. $\dag$: the values are averages of five distinct trials with five different random seeds. $*$: not available as the implementation was very slow. \textbf{Speed} is the relative speed to \baseline{} (larger is faster).}
  \label{tab:bleu-score}
\end{table*}

\section{Learning Curve of Each Model}
\begin{figure*}[ht]
  \begin{subfigure}[b]{0.33\textwidth}
    \centering
  \includegraphics[width=\hsize]{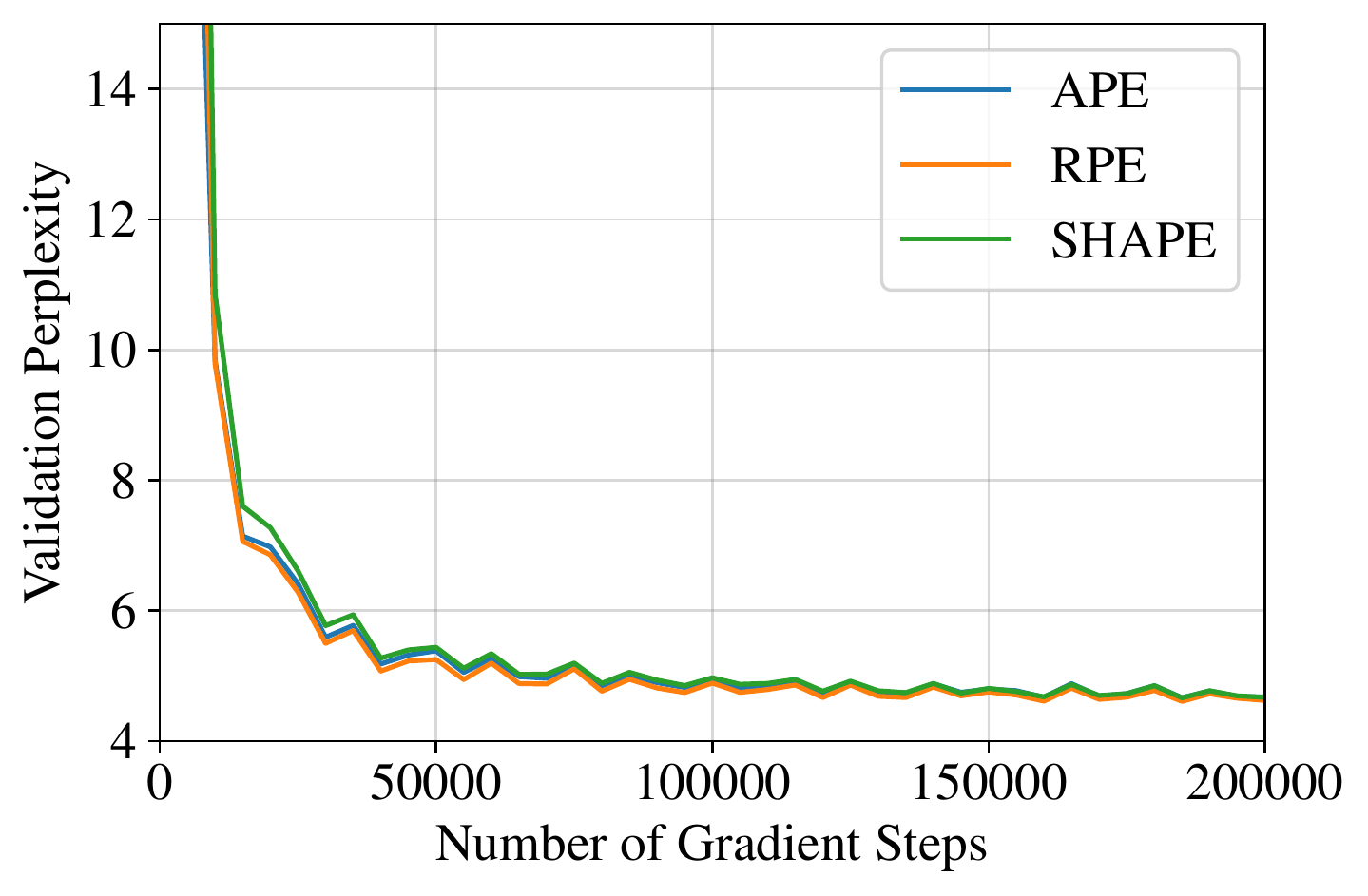}
   \caption{\fulldata{} dataset}
   \label{fig:lr-curve-full-step}
  \end{subfigure}
  \begin{subfigure}[b]{0.33\textwidth}
    \centering
  \includegraphics[width=\hsize]{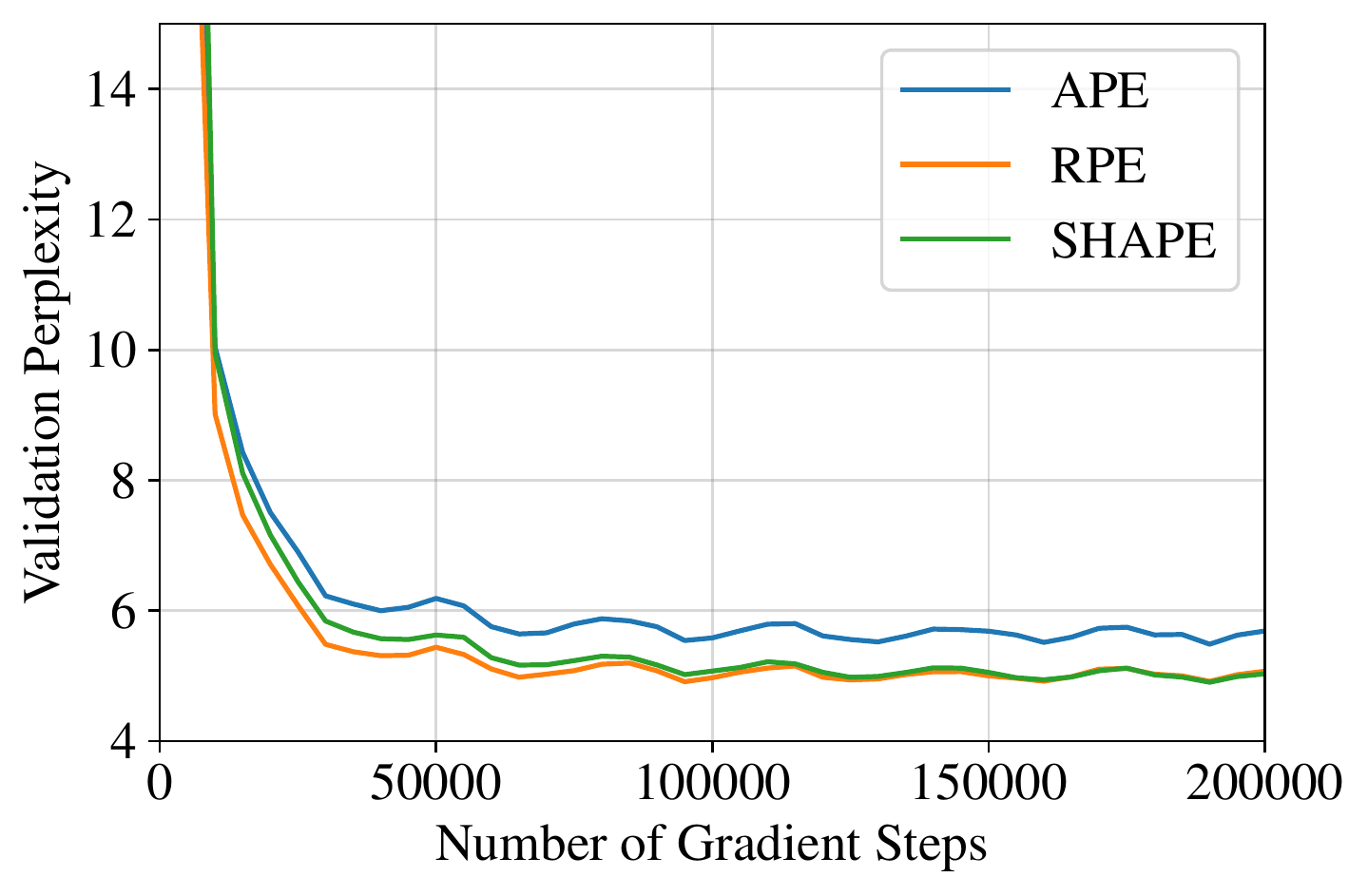}
   \caption{\shortdata{} dataset}
   \label{fig:lr-curve-short-step}
  \end{subfigure}
  \begin{subfigure}[b]{0.33\textwidth}
    \centering
  \includegraphics[width=\hsize]{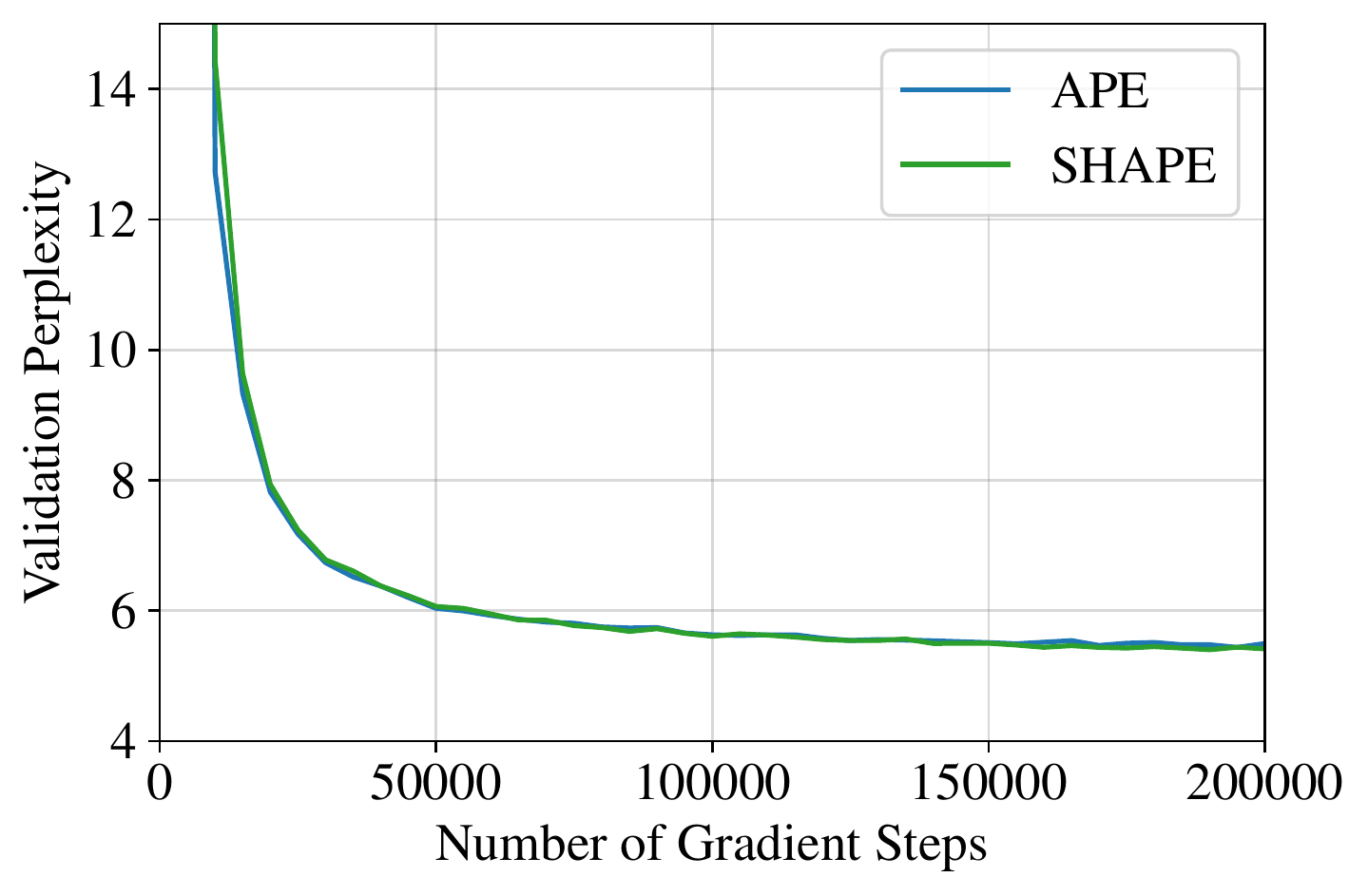}
   \caption{\longdata{} dataset}
   \label{fig:lr-curve-long-step}
  \end{subfigure}
  \caption{Learning curves for each position representation and dataset. We compare the speed of convergence in terms of \textbf{number of gradient steps}.}
   \label{fig:lr-curve-update}
\end{figure*}

\begin{figure*}[ht]
  \begin{subfigure}[b]{0.33\textwidth}
    \centering
  \includegraphics[width=\hsize]{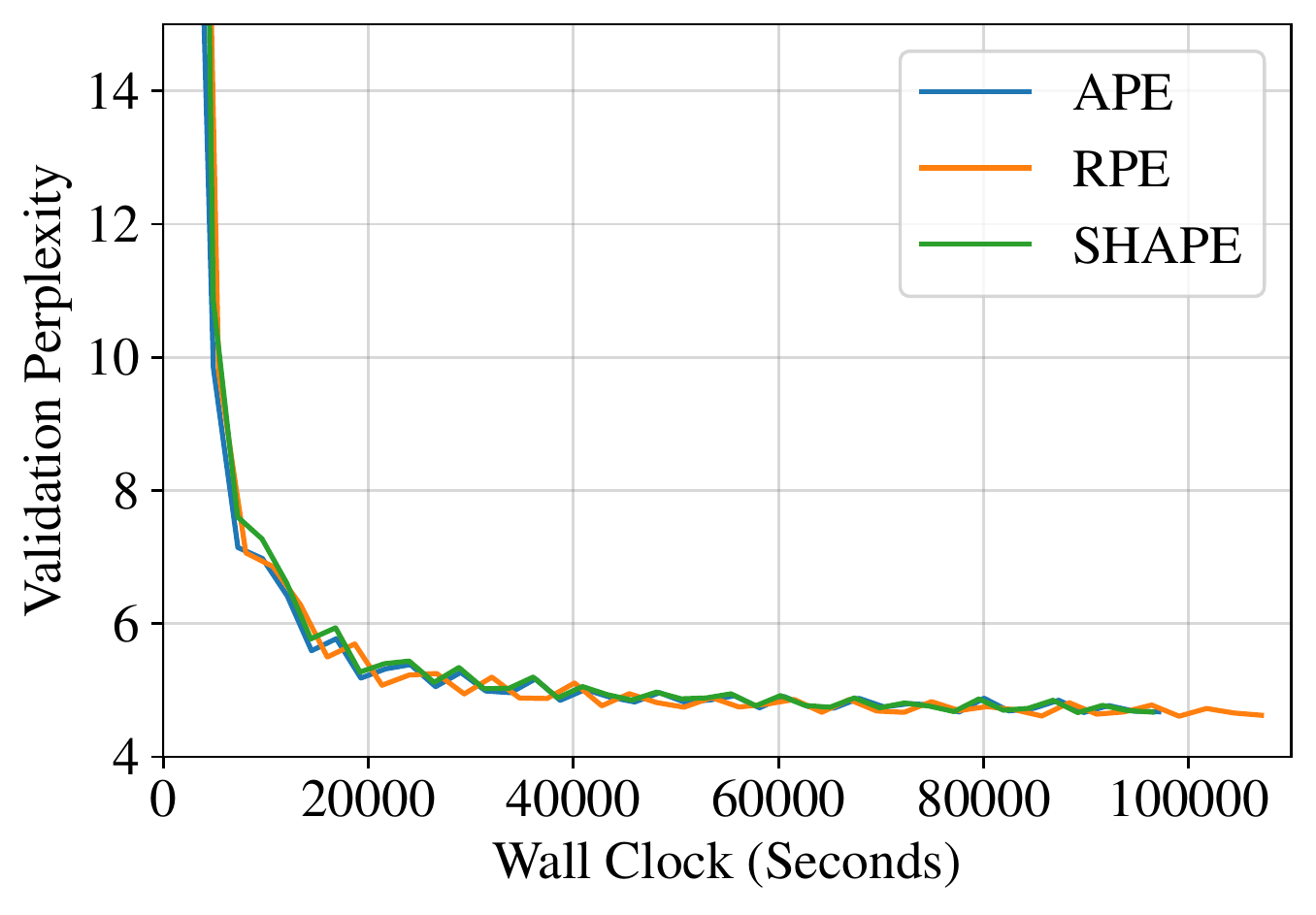}
   \caption{\fulldata{} dataset}
   \label{fig:lr-curve-full-time}
  \end{subfigure}
  \begin{subfigure}[b]{0.33\textwidth}
    \centering
  \includegraphics[width=\hsize]{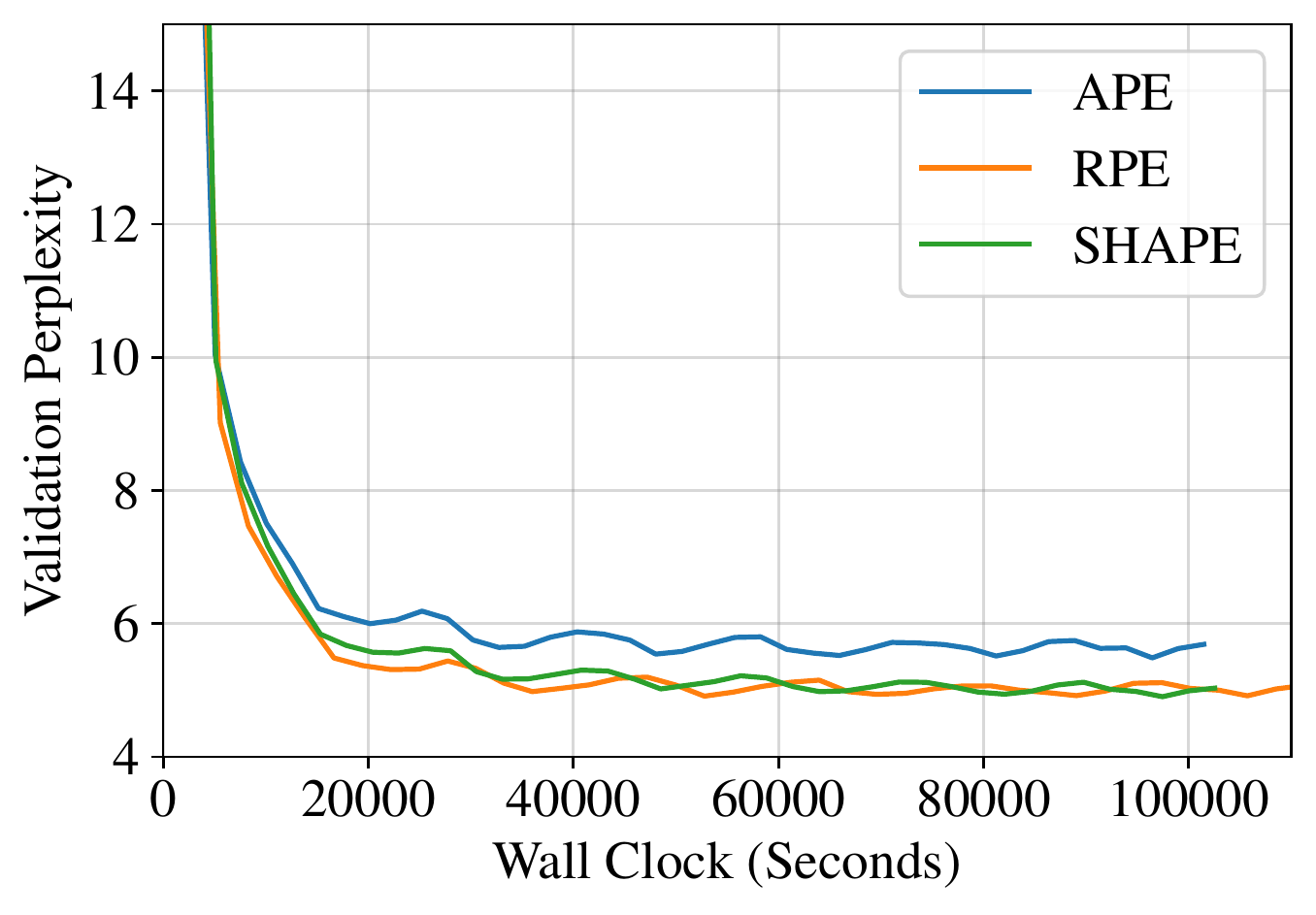}
   \caption{\shortdata{} dataset}
   \label{fig:lr-curve-short-time}
  \end{subfigure}
  \begin{subfigure}[b]{0.33\textwidth}
    \centering
  \includegraphics[width=\hsize]{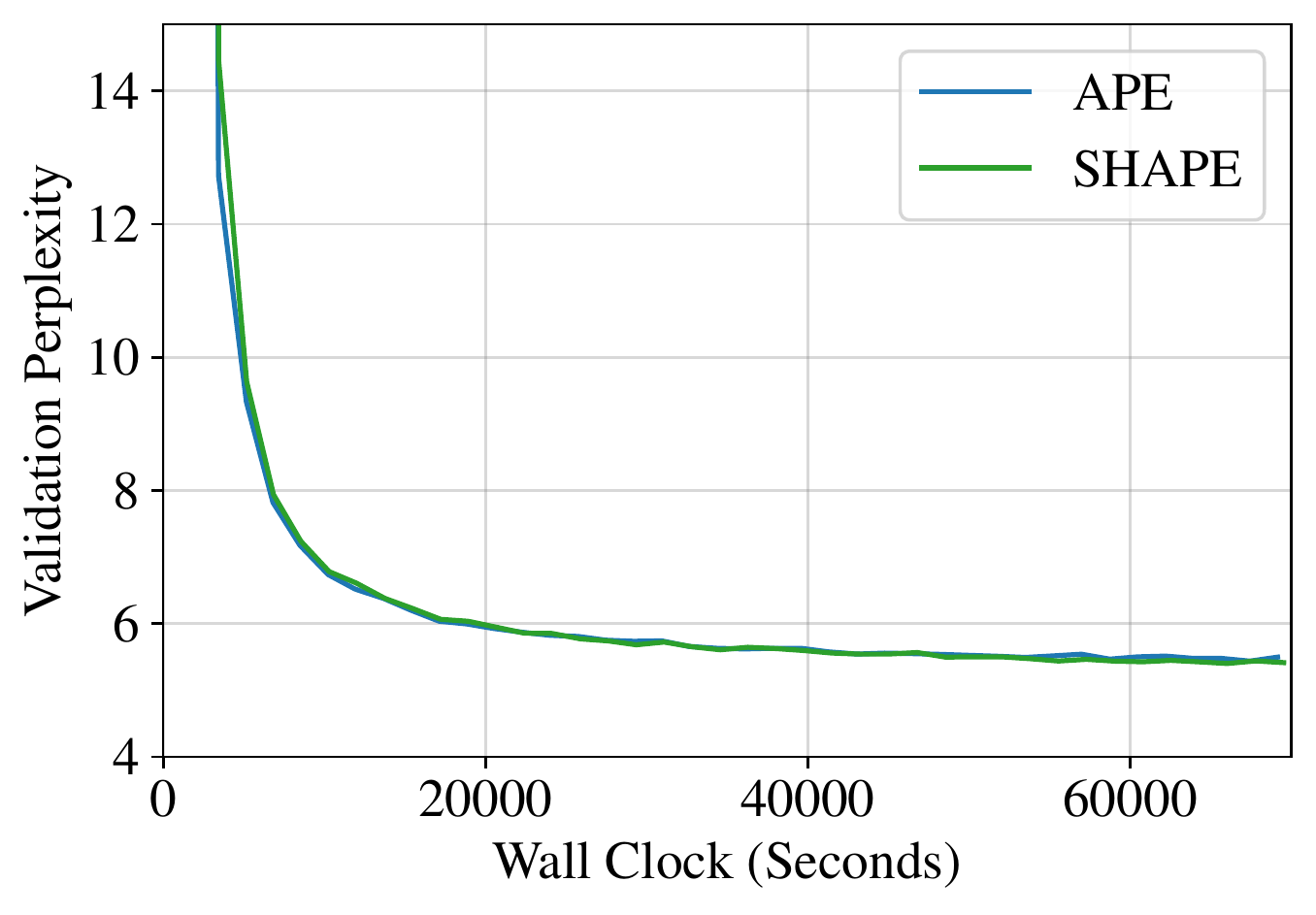}
   \caption{\longdata{} dataset}
   \label{fig:lr-curve-long-time}
  \end{subfigure}
  \caption{Learning curves for each position representation and dataset. We compare the speed of convergence in terms of \textbf{wall clock}.}
   \label{fig:lr-curve-time}
\end{figure*}

We present the learning curve of each model (\baseline{}, \relative{}, \proposed{}) trained on different datasets (\fulldata{}, \shortdata{}, \longdata{}). %
Figures~\ref{fig:lr-curve-update} and \ref{fig:lr-curve-time} present the validation perplexity against the number of gradient steps and wall clock, respectively.
From these figures, we made the following observations:

First, according to Figure~\ref{fig:lr-curve-update}, \textbf{the speed of convergence is similar across the models in terms of the number of gradient steps}.
In other words, in our experiment (Section~\ref{sec:experiment}), we never compare the models whose degree of convergence is different.

Second, Figure~\ref{fig:lr-curve-time} demonstrates that \textbf{\relative{} requires more time to complete the training than \baseline{} and \proposed{} do.}
As explained in Section~\ref{sec:rpe}, RPE causes the computational overhead because it needs to compute attention for relative position embeddings.
The amount of time required to complete the training is presented in Table~\ref{tab:time-required-for-training}.

\begin{table*}[t]
  \centering
  \small
  \begin{tabular}{lllrr}
  \toprule
  \multicolumn{1}{c}{\textbf{Model}} &
    \multicolumn{1}{c}{\textbf{Dataset}} &
    \multicolumn{1}{c}{\textbf{Hardware}} &
    \multicolumn{1}{c}{\textbf{Training Time (sec)}} &
    \multicolumn{1}{c}{\textbf{Number of Parameters}} \\ \midrule
  APE       & \fulldata{}  & DGX-1 & 97,073 & 61M \\
  RPE       & \fulldata{}  & DGX-1 & 107,089& 61M \\
  \proposed{} & \fulldata{}  & DGX-1 & 96,439 & 61M \\ \midrule
  APE       & \shortdata{} & DGX-1 & 101,469& 61M \\
  RPE       & \shortdata{} & DGX-1 & 111,246 & 61M\\
  \proposed{} & \shortdata{} & DGX-1 & 102,535& 61M \\ \midrule
  APE       & \longdata{}  & DGX-2 & 69,148 & 61M \\
  \proposed{} & \longdata{}  & DGX-2 & 69,529 & 61M \\ \bottomrule
  \end{tabular}
  \caption{Training time required to complete 200,000 gradient steps. \relative{} requires more time than \baseline{} and \proposed{} do. Figure~\ref{fig:lr-curve-time} illustrates the corresponding learning curve.}
  \label{tab:time-required-for-training}
\end{table*}

\section{Sanity Check of the Baseline Performance}
\label{sec:fairseq-exp}

Building a strong baseline is essential for trustable results~\citep{denkowski-neubig-2017-stronger}.
To confirm that our baseline model (i.e., Transformer with APE) trained using OpenNMT-py~\citep{klein-etal-2017-opennmt} is strong enough, we compared its performance with that trained on Fairseq~\citep{ott:2019:fairseq}.
Fairseq is another state-of-the-art framework used by winning teams of WMT shared task~\citep{ng:2019:wmt}.
For training on Fairseq, we used the official recipe available in the documentation\footnote{\url{https://github.com/pytorch/fairseq/tree/master/examples/scaling_nmt}}.
The result is presented in Table~\ref{tab:fairseq-result}.
Here, the results are the average of five distinct trials with different random seeds.
From the table, we can confirm that both models can achieve comparable results.

\begin{table*}[ht]
  \small
  \centering
  \begin{tabular}{llccccccc|c}
  \toprule
  \multicolumn{1}{c}{\textbf{Model}} & \multicolumn{1}{c}{\textbf{Implementation}} & \textbf{2010}  & \textbf{2011}  & \textbf{2012}  & \textbf{2013}  & \textbf{2014}  & \textbf{2015}  & \textbf{2016}  & \textbf{Average} \\ \midrule
  APE                  & Fairseq                            & 24.24 & 22.10 & 22.40 & 26.38 & 27.11 & 29.58 & 34.34 & 26.59   \\
  APE                  & OpenNMT-py                         & 24.22 & 21.98 & 22.20 & 26.06 & 26.95 & 29.98 & 34.46 & 26.55   \\ \bottomrule
  \end{tabular}
  \caption{BLEU score on newstest2010-2016. We report average result of five distinct trials with different random seeds.}
  \label{tab:fairseq-result}
\end{table*}

\end{document}